\documentclass[final,5p,times,twocolumn]{elsarticle}

\usepackage{amssymb}
\usepackage{amsmath,amssymb,amsfonts}
\usepackage{algorithmic}
\usepackage{graphicx}
\usepackage{textcomp}
\usepackage{hyperref}
\usepackage{booktabs}
\usepackage{adjustbox}
\usepackage{multirow}
\usepackage{xcolor}
\usepackage{colortbl}
\usepackage{arydshln}
\usepackage{array}
\usepackage{stfloats}
\usepackage{bm}
\usepackage{pifont}
\usepackage{makecell}

\journal{set me later}

\definecolor{DarkGreen}{rgb}{0.2,0.5,0.2} 
\definecolor{mediumgray}{gray}{0.4} 
\newcommand{\etal}{et al.}
\newcommand{\cmark}{\ding{51}} 
\newcommand{\xmark}{\ding{55}} 

\begin{document}
\begin{frontmatter}


\title{Faces of Fairness: Examining Bias in Facial Expression Recognition Datasets and Models} 

\author[]{Mohammad~Mehdi~Hosseini} 
\author[]{Ali~Pourramezan~Fard}
\author[]{Mohammad~H.~Mahoor}

\affiliation{organization={Ritchie School of Engineering and Computer Science, University of Denver},
            city={Denver},
            postcode={80208}, 
            state={CO},
            country={USA}}
            
\begin{abstract}
Automated Facial Expression Recognition (FER), involves two critical aspects: data and model design. Both significantly influence bias and fairness in FER tasks. However, issues related to bias and fairness in FER datasets and models remain underexplored. This study investigates bias and fairness in FER datasets and models. The bias of four common in-the-wild FER datasets, including AffectNet, ExpW, Fer2013, and RAF-DB, is studied. Additionally, this research evaluates the bias and fairness of seven deep models, including three generic CNN models: MobileNet, ResNet, XceptionNet, as well as two popular Transformer-based models: ViT and CLIP, plus two FER-specific state-of-the-art models: POSTER and CEPrompt. Unlike prior studies that examine only limited aspects of bias, our work introduces a unified evaluation framework for FER that integrates five existing and two newly proposed dataset metrics with four fairness criteria for model analysis. We further introduce two new metrics, Conditional-Entropy Bias Index and Concentration Index, designed to quantify conditional dependencies and intra-group data imbalance that existing measures fail to capture. Our results show that all four datasets carry significant demographic bias, most notably in race, with AffectNet exhibiting the highest overall bias and Fer2013 the lowest. At the model level, we find that residual-based CNN architectures (ResNet and XceptionNet) exhibit the lowest overall bias, whereas Transformer-based models (ViT and CLIP) exhibit the highest, despite often achieving comparable or superior accuracy. These findings demonstrate that high predictive accuracy does not guarantee fairness, and that dataset-level and model-level bias must be addressed jointly rather than in isolation.
\end{abstract}



\begin{keyword}
Bias \sep Fairness \sep Models \sep Datasets \sep Facial analysis \sep Affective Computing \sep Facial Expression Recognition
\end{keyword}

\end{frontmatter}
\section{Introduction}
\label{sec:introduction}
Facial expressions serve as a key non-verbal communication channel for humans~\cite{schirmer2024non}, allowing them to express emotions and shape behavior, social interactions, and lifestyle~\cite{dolan2002emotion}. Automated facial expression recognition (FER) has a broad range of applications, including mental health monitoring~\cite{jiang2020facial}, human-computer interaction (HCI)\cite{pantic2003toward}, and surveillance\cite{zhang2018facial}. Deep learning methods have recently achieved significant progress in FER, particularly through the use of Convolutional Neural Networks (CNNs)~\cite{borgalli2025hybrid, zhang2024recognizing}, Transformers~\cite{fatima2025enhanced}, and Large Vision-Language models~\cite{saadi2024leveraging}, as well as interpretable action-unit-driven approaches~\cite{wan2026bipg}. Recent studies have further demonstrated the effectiveness of deep learning-based FER systems across diverse real-world scenarios, highlighting both their performance and limitations under varying conditions~\cite{mejia2023improving}.

Despite advancements in deep learning, bias and fairness issues in FER systems remain unexplored. In a general overview, bias refers to prejudices or favoritism toward groups, while fairness ensures impartiality and avoids disparities~\cite{mehrabi2021survey, barocas2023fairness}. Bias can be examined across two key dimensions: datasets and models~\cite{mehrabi2021survey, suresh2019framework, ferrara2023fairness}. Dataset bias often arises from demographic imbalances, such as disparities in age, gender, and race~\cite{mehrabi2021survey}. Model bias stems from architecture, training process, and unrepresentative datasets~\cite{liang2024linking, caro2020local, dablain2024towards, veeramachaneni2025large}. Bias impacts model performance for underrepresented groups, reduces generalizability, and compromises fairness. Therefore, studying bias in datasets and models is crucial for fairness in FER.

There is a notable research gap in the bias of automated FER, particularly in in-the-wild datasets and novel deep models. In-the-wild datasets introduce biases from variations in illumination conditions, camera resolution, background noise, head pose, and gesture, as well as demographic attributes of age, gender, and race~\cite{xu2020investigating, fard2022ad}. These challenges have also been observed in recent FER systems, where variations in imaging conditions and data imbalance significantly impact model robustness and generalization~\cite{mejia2023improving, barros2022across}. Many studies have focused on limited sources of bias in datasets, such as examining label distribution or one or two demographic attributes~\cite{deuschel2020uncovering, li2020deeper, chen2021understanding, dominguez2024metrics}, where they overlook the multifaceted nature of bias in FER. Additionally, recent advancements in deep architectures, such as Vision Transformers~\cite{dosovitskiy2020image} and Large Vision-Language models~\cite{radford2021learning}, have significantly impacted this field. However, their implications for bias and fairness in FER tasks remain largely unexplored. These shortcomings prompted us to conduct an in-depth investigation into bias and fairness in facial expression recognition.

\begin{figure}[t]
\centering
\includegraphics[width=1.0\columnwidth]{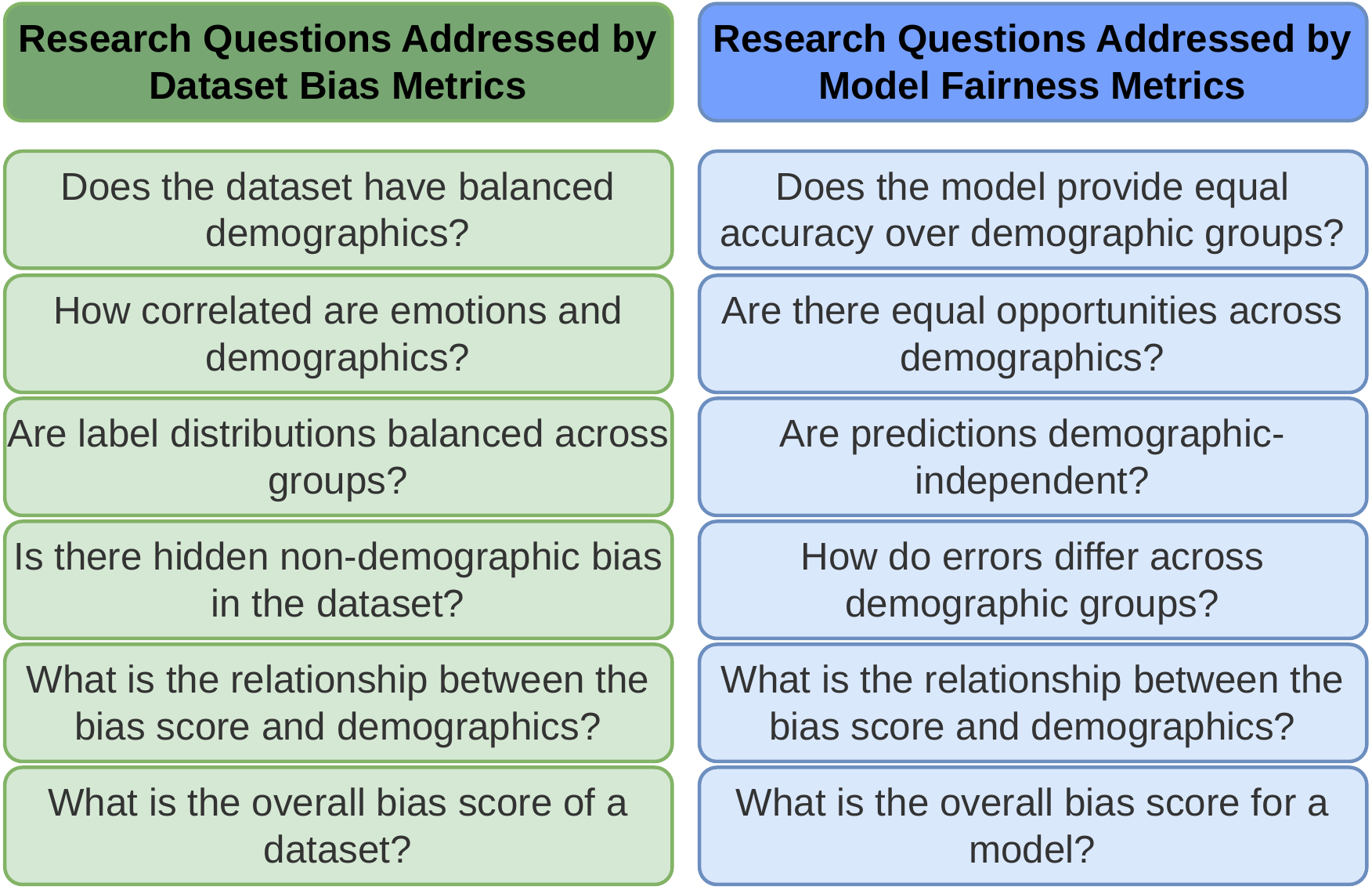}
\caption{The major questions that we analyze in this research aim to study bias and fairness in FER datasets and models.}
\label{fig:proposed_approach}
\end{figure}

This research addresses key gaps in FER fairness by examining overlooked sources of bias in both datasets and deep models. Prior studies typically focus on either dataset imbalance or model-level disparities, but none provide an integrated framework that jointly analyzes both sides using consistent metrics. To bridge this gap, we introduce new FER-specific bias measures, integrate macro- and micro-level co-occurrence analysis with model-based experiments, and evaluate fairness across seven diverse architectures, including FER-specialized and Vision–Language models. Our holistic framework demonstrates that both datasets and models can independently and jointly contribute to biased FER outcomes. Figure~\ref{fig:proposed_approach} summarizes the main research questions of this study.

To assess the impact of datasets, we analyze four commonly used FER datasets: AffectNet~\cite{mollahosseini2017affectnet}, ExpW~\cite{zhang2018facial}, RAF-DB~\cite{li2017reliable}, and Fer2013~\cite{goodfellow2013challenges}. The bias in these datasets is evaluated using seven statistical metrics that measure diversity, uncertainty, and dependence in the data. We also investigate bias through analyses of single and joint probability distributions over the labels and demographic attributes of age, gender, and race. In addition, we examine inherent bias by conducting two model-based experiments: (1) identifying which dataset each sample originates from, and (2) evaluating how well models trained on three datasets generalize to a held-out dataset.

We also study bias in models by evaluating seven popular deep models in FER. Specifically, we select three CNN models: MobileNetV1~\cite{howard2017mobilenets}, ResNet-50~\cite{he2016deep}, and XceptionNet~\cite{chollet2017xception}; two Transformer-based models: ViT~\cite{dosovitskiy2020image} and CLIP~\cite{radford2021learning}; and two state-of-the-art FER-specialized models: POSTER~\cite{zheng2023poster} and CEPrompt~\cite{zhou2024ceprompt}. We assess bias in these models using four key metrics that quantify how predictions or errors vary across demographic groups, capturing disparities in outcomes, opportunities, and relative treatment and evaluating overall fairness. By measuring multiple aspects of bias, these metrics provide a comprehensive evaluation across different attributes, allowing a fair and thorough comparison of the models' bias scores.

The main contributions of this paper are:
\begin{itemize}
    \item A unified framework for dataset- and model-level bias: We propose a unified pipeline that integrates seven dataset metrics, multi-level co-occurrence analysis, dataset-recognition and leave-one-dataset-out experiments, together with four model fairness criteria.
    \item New dataset bias metrics: We develop two new bias metrics, Conditional-Entropy Bias Index (CEBI) and Concentration Index (CI), that quantify conditional label–attribute dependencies and intra-group imbalance. These metrics address limitations of prior dataset-diversity measures and provide a more informative foundation for analyzing FER dataset bias.
     \item Cross-architecture fairness benchmark: We evaluate seven representative models (CNNs, Transformers, LVLMs, and FER-specialized models), providing a reproducible benchmark and revealing insights into architectural sources of bias.
     \item Comprehensive joint interpretation: We analyze how dataset bias may propagate into model bias, offering an integrated perspective that remains relatively underexplored in prior FER fairness studies.
\end{itemize}
Our implementation codes and experiments are available at \href{https://github.com/MMHosseini/bias_in_FER}{https://github.com/MMHosseini/bias\_in\_FER}.

In the remainder of this paper, Sec.~\ref{sec:literature_review} provides a review of the literature on bias and fairness in both datasets and algorithms. Next, Sec.~\ref{sec:methodology} outlines the metrics used to evaluate bias, and Sec.~\ref{sec:experimental_results} discusses bias scores for datasets and models. Finally, Sec.~\ref{sec:discussion_and_future_works} concludes the paper and highlights several open research directions that build on this study.


\section{Literature Review}
\label{sec:literature_review}
Bias in machine learning can be classified in various ways. Pagano~\etal~\cite{pagano2023bias} introduced three types of bias: pre-existing bias, technical bias, and emergent bias. Mehrabi~\etal~\cite{mehrabi2021survey} provided a broader taxonomy includes historical, representation, measurement, aggregation, and evaluation biases. Understanding the sources of bias is essential to effectively address its challenges. Bias in FER can originate from the datasets, the models used for learning, the evaluation protocols, and the deployment conditions. The primary sources of bias in FER include imbalanced demographic representation, data collection and sampling artifacts, cultural variation in expression, subjectivity in the annotation process, limitations in model architectures and objective functions, shortcomings in evaluation methodologies, and deployment-related domain shifts. Developing fair and unbiased deep methods for FER potentially ensures equitable treatment across diverse demographic groups~\cite{amigo2023mitigating}. Thus, dealing with bias in FER has recently raised attention from researchers. 
In the following, we review recent studies on bias in FER datasets and models.

\subsection{Bias in FER Datasets} 
\label{sec:literature_review:bias_in_fer_datasets}
Bias in datasets refers to systematic skews or unfairness in data that can distort model training and predictions~\cite{bernhardt2022potential}. Several taxonomies have been proposed to categorize dataset bias~\cite{fabbrizzi2022survey}. One common distinction is between representational bias and stereotypical bias\cite{dominguez2024metrics, dominguez2022gender, dominguez2022assessing}. The former targets demographic diversity, while the latter examines correlations between subgroups and labels. Another categorization separates intrinsic and extrinsic biases\cite{mavadati2013disfa, srinivas2019face, singh2022anatomizing}: intrinsic bias arises from the data collection process and demographic or behavioral factors, whereas extrinsic bias stems from external conditions such as camera resolution, illumination, background noise, or annotator subjectivity. Additional dimensions have also been identified, with Cheong~\etal~\cite{cheong2023causal} distinguishing among data collection bias, labeling bias, and contextual bias, and Jones~\etal~\cite{jones2024causal} categorizing bias into prevalence disparities, presentation disparities, and annotation disparities. Despite these diverse perspectives, there is strong consensus that demographic attributes, particularly age, gender, and race, remain the most influential sources of dataset bias and warrant the most attention.

\begin{figure}[t!]
\centering
\includegraphics[width=1.0\columnwidth]{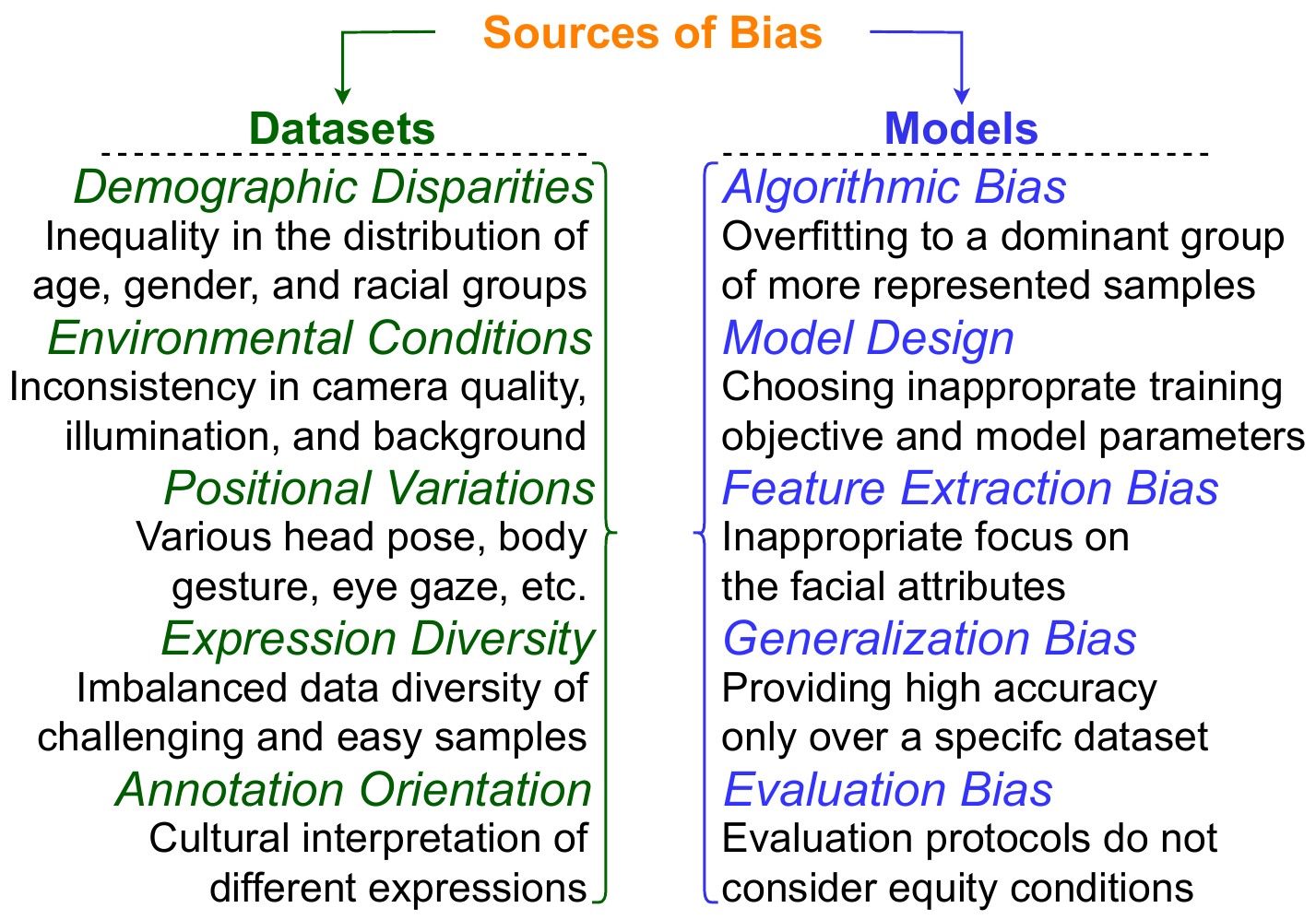}
\caption{Bias can originate from datasets and models. In datasets, prominent sources of bias include issues in data collection, such as demographic disparities, variations in illumination and lighting conditions, gestures, head poses, and cultural differences in emotional interpretation. In models, key bias sources include architecture design, training parameters, overfitting to specific demographic groups, and the selection of evaluation metrics.}
\label{fig:bias_sources_in_fer}
\end{figure}

In the matter of in-the-wild FER datasets analysis, Dominguez~\etal~\cite{dominguez2024metrics} found that while they are diverse in terms of demographic representation, they still exhibit weaknesses under stereotypical bias metrics. Li~\etal~\cite{li2020deeper} analyzed intrinsic bias across seven FER datasets, identifying cultural differences and data collection conditions as the major contributions. Huber~\etal~\cite{huber2024bias} showed that synthetic facial datasets replicate the bias patterns present in the authentic datasets used for training. Cheong~\etal~\cite{cheong2023causal} employed directed acyclic graphs to examine bias sources, highlighting gender and age as the most influential factors in emotion labeling. Collectively, these studies emphasize the critical role of dataset analysis in FER tasks, particularly the impact of demographic attributes on label distributions~\cite{georgescu2019local}. Similar observations have been reported in recent FER studies, where imbalanced expression distributions and environmental variations were shown to influence model performance and bias characteristics~\cite{barros2022across}.

Although establishing a taxonomy of dataset bias is important, understanding the underlying sources of bias provides guidance for mitigation. Based on our study, we categorize sources of bias into five main classes: 1) demographic disparities, referring to imbalances in age, gender, and racial groups that can affect model fairness across different populations; 2) environmental conditions, capturing inconsistencies in illumination, background scenes, and camera quality that may influence visual perception; 3) positional variations, including differences in body gestures, eye-gaze, head-pose, and clothing, which can alter the appearance of expressions; 4) expression diversity, reflecting imbalances between labels, and also easily recognizable and more subtle or challenging expressions that can bias model learning; and 5) annotation orientation, addressing cultural and subjective differences in labeling, which may introduce inconsistencies in ground truth data. A visual summary of these sources of bias is provided in Fig.~\ref{fig:bias_sources_in_fer}.

\subsection{Bias in FER Algorithms} 
\label{sec:literature_review:bias_in_fer_algorithms}
Model bias is a systematic error in decision-making that produces unfair outcomes~\cite{ferrara2023fairness}. Another view describes it as biased results caused by inadequate model specifications~\cite{akter2022algorithmic}. Several taxonomies of bias in ML models have been proposed. Mehrabi~\etal~\cite{mehrabi2021survey} identify algorithmic bias, evaluation bias, and optimization bias as key forms: algorithmic bias reflects the model’s assumptions, evaluation bias stems from benchmarks favoring certain groups, and optimization bias arises when objectives or loss functions benefit majority groups. Binns~\etal~\cite{binns2023legal}, instead, classify bias as direct, indirect, and structural groups. Direct bias occurs when algorithms deliberately treat groups unfairly to boost accuracy, indirect bias emerges from accidental errors when models fail to capture features, and structural bias results from organizational inequalities shaping algorithmic behavior.

Recent studies of bias in ML algorithms reveal that unfair outcomes in FER often persist even when data issues are addressed, suggesting that models themselves introduce bias. Dooley~\etal~\cite{dooley2024rethinking} showed that bias can stem directly from network architecture, while Abdullah~\etal~\cite{abdullah2024effects} found that attention mechanisms allocate less focus to darker-skinned faces. Xu~\etal~\cite{xu2020investigating} reported that data augmentation offers little improvement when the model is inherently biased, and Amigo~\etal~\cite{amigo2023mitigating} demonstrated that models can reproduce bias across datasets despite debiasing strategies. Dominguez~\etal~\cite{dominguez2024less} further argued that stereotypical biases are intrinsic to models, regardless of data balance. Together, these findings highlight that algorithmic bias is deeply embedded in model architectures and learning mechanisms, not just in the data. However, these findings do not imply that data bias is inconsequential. On the contrary, biases in the data can still significantly influence and amplify model bias. Recent deep learning-based FER systems further confirm that architectural design and training strategies can significantly influence performance disparities across different input conditions and demographic variations~\cite{han2020toward}.

Similar to dataset bias, we identify five main sources of bias in ML models: (1) algorithmic bias, which occurs when the model overfits to a dominant group in the training data; (2) model design bias, arising from choices in architecture, objective functions, loss functions, or training strategies; (3) feature extraction bias, when the model places undue emphasis on certain facial attributes; (4) generalization bias, where the model performs well only on a specific dataset but fails to transfer to others; and (5) evaluation bias, which stems from benchmarks or metrics that overlook fairness considerations (see Fig.~\ref{fig:bias_sources_in_fer}).

\subsection{Perspectives}
Bias and fairness in FER are critical for preventing discrimination, ensuring equitable performance across demographic groups, and upholding ethical standards in affective computing systems. Because modern FER models rely heavily on in-the-wild datasets, it is essential to examine these datasets to uncover and understand the diverse sources of bias they contain. While previous studies have explored aspects of demographic imbalance or label distribution, a more systematic and metric-rich examination is needed to reveal deeper, previously undiscovered forms of dataset bias. At the same time, rapid advancements in model architectures, from CNNs to Transformers and Large Vision-Language models, have significantly improved FER accuracy \cite{rodrigo2024comprehensive, aldahoul2024exploring, zhao2024enhancing, zhang2024recognizing}. Despite these improvements, questions of fairness and algorithmic bias in modern FER models remain underexplored.

Existing fairness research in FER tends to analyze isolated components, focusing either on demographic skew in datasets or on disparities in model predictions. To the best of our knowledge, no prior work integrates new dataset-bias metrics, multi-level co-occurrence analyses, model-based dataset generalization tests, and fairness evaluation of contemporary architectures within a single unified framework. This gap is significant because fully understanding bias in FER requires examining how dataset imbalance, label dependencies, and hidden non-demographic artifacts propagate into model behavior. Our work addresses this need by providing the first end-to-end FER bias analysis capable of jointly quantifying dataset diversity, conditional dependencies, latent dataset-specific features, and architecture-sensitive model bias.

In this paper, we analyze four widely used FER datasets (AffectNet, ExpW, Fer2013, and RAF-DB) using seven complementary statistical and information-theoretic metrics, combined with co-occurrence-based and model-based evaluations. In parallel, we assess seven representative FER models, including five general-purpose architectures (MobileNet, ResNet, XceptionNet, ViT, and CLIP) and two FER-specialized models (POSTER and CEPrompt), to examine how bias manifests across architectural families and performance levels. Taken together, this constitutes the first systematic evaluation that simultaneously investigates dataset-level bias and model-level fairness in FER, enabling a more holistic understanding of how bias originates, how it propagates, and how it evolves with advancements in deep learning.

\section{Methodology} 
\label{sec:methodology}
Bias and fairness are related but distinct concepts (see~\cite{verma2018fairness}). In this study, we operationalize bias as any measurable imbalance or dependency in the dataset, quantified using the five existing and two newly proposed metrics described in Sec.~\ref{sec:methodology:data_analysis}. Fairness is also operationalized in terms of model behavior and evaluate it using four established metrics that measure whether prediction outcomes are distributed equitably across demographic groups (see Sec.~\ref{sec:methodology:model_analysis}). These definitions form the basis of our methodology and allow us to systematically assess dataset-level bias and model-level fairness in FER. Table~\ref{tbl:taxonomy} presents the taxonomy of our equations.

\begin{table}[t!] 
\caption{Taxonomy of the equations used in this paper.}
\label{tbl:taxonomy}
\centering
\small
\resizebox{0.48\textwidth}{!}
{{
\begin{tabular}{lll}
\toprule
\textbf{Symbol} & \textbf{Meaning} & \textbf{Example} \\
\midrule
$A, B, C$   & Symbol of each attribute & $A$: Age, $B$: Gender\\
$a, b, c$   & Specific demographic group & $a$: Man, $b$: Woman\\
$Y$         & Expression set (labels) & - \\
$y$         & Specific expression value & $y$: Happy\\
$H$         & Entropy & $H(Y)$: Entropy of labels\\
$MET$       & Models' bias metrics & -\\
$ATT$       & Attributes (age, gender, race) & -\\
$p$         & Probability & -\\
$n$         & Number of labels & 7 in our study \\
$m$         & Number of metrics & - \\
$d$         & Number of attributes & 3 in our study \\
$k$         & Number of demographic groups & 4 age groups\\
\bottomrule
\end{tabular}
}}
\end{table}
\begin{table*}[t!]
\centering
\caption{Characteristics of the seven data analysis metrics used to measure dataset bias, including Wasserstein Distance (WD), Jensen–Shannon Divergence
(JSD), Normalized Shannon Entropy (NSE), Normalized Label Skewness (NLS), Group Normalized Mutual Information (GNMI), Conditional-Entropy Bias Index (CEBI), and Concentration Index (CI).}
\label{tbl:data_metrics_summary}
\setlength{\tabcolsep}{18pt} 
\resizebox{1.0\textwidth}{!}
{
\renewcommand{\arraystretch}{1.5}
\begin{tabular}{m{0.9cm} m{4.2cm} m{4.2cm} m{7.2cm}}
\toprule
\textbf{Metric} & \textbf{Measures} & \textbf{Interpretation} & \textbf{Example}\\ 
\midrule
WD & Distributional difference of groups; mass transport cost & 0: identical distributions; 1: maximally different &  Male and Female groups have different probability distributions for the Happy expression \\ 

JSD & Difference in probabilistic information content across groups & 0: identical distribution; 1: disjoint distributions & Asian and Black groups show highly different expression distributions, leading to high divergence \\

NSE & Overall label diversity within each group & Low entropy: dominance; high entropy: balanced diversity & The race attribute shows low entropy for Angry, where one race accounts for most Angry labels \\ 

NLS & Directional asymmetry of label distribution within groups & High: strong over- or under-representation of certain labels & The Female group’s expression distribution is skewed toward Neutral compared to the overall mean \\ 

GNMI & Statistical dependency between label and attribute & High: label highly predictable from demographic variable & Membership in the Black racial group strongly predicts the Surprise label, indicating high mutual information \\

CEBI & Dependence of label on demographic attribute & High: strong dependence of labels on groups & Knowing $age\mathord{=}[0\mathord{\sim}15]$ substantially reduces uncertainty about the Happy expression\\ 

CI & Evenness of label distribution within each group & High: a few labels dominate within a group & Within the Male group, Neutral and Happy dominate most samples, creating an imbalanced internal distribution \\ 
\bottomrule
\end{tabular}
}
\end{table*}

\subsection{Data Analysis} 
\label{sec:methodology:data_analysis}
Bias in data manifests in ways, such as camera inconsistencies, background variations, body posture differences, and, most notably, demographic imbalances. Among these, demographic attributes (age, gender, and race) emerge as the most significant contributors to bias in FER datasets. As a result, these attributes, along with label distribution, warrant the most attention in this study to address bias and promote fairness.

To evaluate demographic bias in the datasets, we employ Wasserstein Distance (WD)~\cite{miroshnikov2022wasserstein}, Jensen–Shannon Divergence (JSD)~\cite{lin2002divergence}, Normalized Shannon Entropy (NSE)~\cite{shannon1948mathematical}, Normalized Label Skewness (NLS), and Group Normalized Mutual Information (GNMI)~\cite{strehl2002cluster}, as well as two new metrics introduced in this paper, Conditional-Entropy Bias Index (CEBI), Concentration Index (CI). Unlike the metrics in~\cite{dominguez2024metrics}, these measures directly capture distributional differences and conditional dependencies that influence machine-learning models. WD and JSD quantify divergence in label distributions across demographic groups, revealing subtle shifts that broad diversity scores may miss. GNMI and the new metric CEBI measure information gain about labels from demographic attributes, making them sensitive to conditional or interaction-based bias. NSE and the new NLS metric offer normalized, interpretable indicators of entropy and skewness, enabling clear cross-dataset comparisons. We customized all the metrics in a way to work as a bias score in the range of $[0, 1]$, where 0 means no bias and 1 shows the maximum bias.

Because our metrics draw on probability and information theory, they relate more directly to model training and predictive fairness than the ecological variety measures in~\cite{dominguez2024metrics}.  
This connection clarifies how demographic imbalance affects classifier performance and guides mitigation strategies.  
While classical diversity metrics help describe overall variety, the selected measures provide a complementary, machine-learning-focused view of dataset bias.  
Next, we present details for each metric. Table~\ref{tbl:data_metrics_summary} summarizes the metrics we utilize for the data analysis.
\\
\textbf{Wasserstein Distance (WD)}~\cite{miroshnikov2022wasserstein} is a metric sensitive to imbalance magnitude and class structure. For example, in a similar distribution of data for two attributes Male and Female, Wasserstein Distance is equal to 0, while for two skewed and dissimilar distribution it gives a high value. This metric captures how much "probability mass" should be shifted to reach to an equal data distribution: 
\begin{equation}
\label{eq:wasserstein_distance}
\begin{adjustbox}{width=0.86\columnwidth}
$\begin{aligned}
WD = \frac{2}{k(k-1)} \sum_{a, b \in A}\frac{1}{n} \sum_{i=1}^{n} \lvert p(y_i|A=a) - p(y_i|A=b) \rvert.
\end{aligned}$
\end{adjustbox} 
\end{equation}
\textbf{Jensen–Shannon Divergence (JSD)}~\cite{lin2002divergence} quantifies the similarity between two probability distributions. While WD quantifies bias by measuring the minimal effort required to transport one group’s distribution to another, Jensen–Shannon Divergence captures differences in their probabilistic information content without accounting for geometric distance. Our JSD metric is equal to 0 when two distributions are similar, and 1 when they are disjoint. In FER it determines how different the labels are distributed across demographic groups: 
\begin{equation}
\label{eq:jensen_shannon_divergence}
\begin{adjustbox}{width=0.83\columnwidth}
$\begin{aligned}
JSD = \frac{2}{k(k-1)} \sum_{a, b \in A} \frac{1}{2n} \sum_{i=1}^{n} \left(H(a, m) + H(b, m)\right),\\
H(a, m) = p(y_i|A=a) \log \frac{p(y_i|A=a)}{m(y_i)},  \quad\quad\quad\\
m(y_i) = \frac{p(y_i|A=a) + p(y_i|A=b)}{2}.\quad\quad\quad\quad
\end{aligned}$
\end{adjustbox} 
\end{equation}
\textbf{Normalized Shannon Entropy (NSE)}, derived from~\cite{shannon1948mathematical}, summarizes the entire demographic distribution into a single value without requiring pairwise group comparisons. NSE measures the overall label diversity within a group, providing a scale-invariant view of internal balance. For instance, it can evaluate how fairly the Happy expression is distributed across various racial groups; if it is biased toward an overrepresented group, NSE assigns a higher bias score. This metric simultaneously captures both the richness and evenness of group representation: 
\begin{equation}
\label{eq:normalized_shannon_entropy}
\begin{adjustbox}{width=0.66\columnwidth}
$\begin{aligned}
NSE = \frac{1}{k} \sum_{a \in A} p(A=a) \left|1 - \frac{H(Y \mid a)}{\log n} \right|.
\end{aligned}$
\end{adjustbox} 
\end{equation}
\textbf{Normalized Label Skewness (NLS)} is a normalized version of the skewness~\cite{doane2011measuring} metric and quantifies the asymmetry of class label distributions within each demographic group. For example, for the Male group, it measures how much each expression’s distribution deviates from the overall average. While Normalized Shannon Entropy (NSE) captures overall label diversity and balance, NLS highlights the directional imbalance of over- or underrepresented classes within a group. Large positive or negative values indicate strong bias, so NLS is normalized to the range [0, 1] for comparability:
\begin{equation}
\label{eq:normalized_label_skewness}
\begin{adjustbox}{width=0.90\columnwidth}
$\begin{aligned}
NLS = \frac{1}{k} \sum_{a \in A}\frac{|S|}{1+|S|}, \quad\quad S = \frac{1}{n} \sum_{i=1}^{n} \left[\frac{p(y_i|A=a) - \mu}{\sigma}\right]^3.
\end{aligned}$
\end{adjustbox} 
\end{equation}
\textbf{Group-Normalized Mutual Information (GNMI)}~\cite{strehl2002cluster} measures non-linear statistical dependencies between labels and demographic groups. It quantifies how much information about a label can be inferred from the group membership. For example, it evaluates the relationship between the Black racial group and the Happy expression, capturing how much knowing the demographic group reduces uncertainty about the label. While CEBI measures conditional label uncertainty given the group, GNMI normalizes this dependency for scale-independent comparison. High GNMI values indicate that the label is highly predictable from the demographic attribute:
\begin{equation}
\label{eq:group_normalized_mutual_information}
\begin{adjustbox}{width=0.63\columnwidth}
$\begin{aligned}
GNMI(Y,A) = \frac{I(Y;A)}{\sqrt{H(Y)H(A)}}, \quad\quad\quad \\ 
I(Y;A) = \sum_{a \in A}\sum_{y \in Y}p(y,a) \log\frac{p(y, a)}{p(y)p(a)},\quad
\end{aligned}$
\end{adjustbox} 
\end{equation}

Existing distribution-based fairness and imbalance metrics capture only partial aspects of dataset bias. Metrics such as WD and JSD mainly quantify inter-group distribution differences, while entropy- and mutual-information-based measures focus on global uncertainty or dependency. However, these metrics do not explicitly capture conditional label dependence within demographic groups or the degree of intra-group concentration imbalance. To address these limitations, we introduce the Conditional-Entropy Bias Index (CEBI) and Concentration Index (CI), which provide complementary views of dataset bias. CEBI measures how strongly demographic attributes reduce uncertainty in expression labels, while CI quantifies the concentration of label distributions within demographic groups independently of inter-group divergence. Here, we introduce these two metrics and in ~\ref{app:appendix_a} discuss them in detail.
\\
\textbf{Conditional-Entropy Bias Index (CEBI)}, proposed in this paper, is an entropy-based metric that we developed (inspired by~\cite{fischer2020conditional}) to measure how an emotion class depends on a specific attribute in a group. For example, how the Male gender influences the Happy expression. A low CEBI means that the attribute strongly determines the label. To provide a consistent scale in which 0 indicates no bias and 1 indicates maximum bias, we normalized its equation and then reversed by subtracting it from 1:
\begin{equation}
\label{eq:conditional_entropy_bias_index}
\begin{adjustbox}{width=0.66\columnwidth}
$\begin{aligned}
CEBI = \frac{1}{k}\sum_{a\in A} \max\!\left(0,\; 1 - \frac{H(Y\mid a)}{H(Y)}\right),
\end{aligned}$
\end{adjustbox} 
\end{equation}
\\Unlike GNMI, which measures symmetric statistical dependence between variables, CEBI explicitly quantifies the conditional reduction in label uncertainty given demographic attributes, making it more sensitive to localized dependency patterns within specific groups.
\\
\textbf{Concentration Index (CI)} is a new metric that we developed, based on the Gini coefficient~\cite{gini1912variability} and Herfindahl–Hirschman Index~\cite{hirschman1980national}, to quantify the imbalance data distribution across the dataset in the range of [0, 1]. Unlike WD and JSD, which assess differences between group distributions, Concentration Index measures how concentrated or uneven the label distribution is within each group, capturing intra-group imbalance rather than inter-group distance. For example, it evaluates how even the expressions are distributed on Male gender, independent of the Female gender. A higher CI reveals that only a few demographic groups dominate the samples, making it a clear indicator of representational bias:
\begin{equation}
\label{eq:concentration_index}
\begin{adjustbox}{width=0.78\columnwidth}
$\begin{aligned}
CI =  \frac{1}{k} \sum_{a \in A} \frac{\left|\|\mathbf{p}\|_2^2 - \tfrac{1}{n}\right|}{1 - \tfrac{1}{n}}, \quad\quad\quad\|\mathbf{p}\|_2^2 = \sum_{y \in Y} p(y)^2.
\end{aligned}$
\end{adjustbox} 
\end{equation}
\\
While WD and JSD compare distributions across groups, CI focuses on imbalance within individual groups and therefore captures concentration effects that may remain hidden under pairwise divergence measures.

The \textbf{Macro-Level Co-occurrence Bias} and \textbf{Micro-Level Co-occurrence Bias} metrics examine the frequency between an attribute and a class label. Macro-Level Co-occurrence Bias focuses on the overall relationship between an attribute and the data distribution, such as the connection between the $age$ attribute and the $expression$ labels. On the other hand, Micro-Level Co-occurrence Bias delves into analyzing the connection between a particular group and a class label. For example, the relationship between the group $age\mathord{=}[0\mathord{\sim}15]$ and $expression\mathord{=}Happy$.

To investigate Macro-Level Co-occurrence Bias, we extracted the probability distributions of expressions and attributes in each dataset, denoted by $p(Y\mathord{=}y)$ and $p(A\mathord{=}a)$, respectively. To study Micro-Level Co-occurrence Bias, given that each expression class is associated with three attributes, i.e. age, gender, and race, we explored the data distribution under single, double and triplet conditions, denoted by $p(Y\mathord{=}y|A\mathord{=}a)$, $p(Y\mathord{=}y|A\mathord{=}a, B\mathord{=}b)$ and $p(Y\mathord{=}y|A\mathord{=}a, B\mathord{=}b, C\mathord{=}c)$, respectively.

We also evaluated non-demographic biases in each dataset. To do so, we designed an experiment where a model is trained to recognize the datasets. In this experiment, each image sample was labeled with its corresponding dataset name, for both the training and validation sets. Then, the model was trained to predict the dataset of each image. High accuracy in this experiment indicates higher non-demographic bias, such as those related to illumination, background noise, camera resolution, gesture, eye gaze, head pose, hair color, and any other hidden bias in the datasets.

In addition, we conducted another experiment to evaluate bias in datasets. In this leave-one-dataset-out experiment, we trained a deep model for facial expression recognition, using all except one dataset, which was left out for test. By assessing the model's performance on the excluded dataset, we aimed to identify potential biases within the dataset. In this experiment, lower accuracy suggests higher bias in the datasets. To ensure a fair comparison, we trained the model on the training sets of all datasets except one, validated its accuracy using their respective validation sets, and tested the bias using the validation set of the excluded (left-out) dataset. If the results of the validation and test experiments show similar accuracies, the model is considered unbiased. Any discrepancy indicates that the dataset suffers from bias. 

To summarize, we utilized seven mathematical data analysis metrics and designed two model-based experiments, including a dataset recognition task and a leave-one-dataset-out task, to assess bias within the datasets.

\begin{table*}[t!]
\centering
\caption{Characteristics of the fairness metrics used to evaluate model bias, including Equal Odds (Eq-Od), Equal Opportunity (Eq-Op), Demographic Parity (De-Pa), and Treatment Equality (Tr-Eq).}
\label{tbl:model_metrics_summary}
\setlength{\tabcolsep}{18pt} 
\resizebox{1.0\textwidth}{!}
{
\renewcommand{\arraystretch}{1.4}
\begin{tabular}{m{1.2cm} m{4.2cm} m{4.8cm} m{6.2cm}}
\toprule
\textbf{Metric} &
\textbf{Measures} &
\textbf{Interpretation} &
\textbf{Example} \\
\midrule
Eq-Od & TP and FP rates of protected vs. unprotected groups & Violations indicate unequal error distribution across groups & Male and Female groups show different TP and FP rates for the Happy expression\\
Eq-Op & TP rates across groups & Measures fairness in detecting the correct positive class & The TP rate for Sad is lower for the Black group than for the White group\\
De-Pa & Overall prediction rates across groups & Ensures predictions are not influenced by sensitive attributes & The model predicts Neutral more frequently for females than males\\
Tr-Eq & FN to FP ratios across groups & High disparity indicates unequal tolerance of mistakes across groups & The FN/FP ratio for Angry is  higher for the Asian group than for the Hispanic\\
\bottomrule
\end{tabular}
}
\end{table*}

\subsection{Model Analysis} 
\label{sec:methodology:model_analysis}
A variety of general metrics have been proposed to evaluate fairness in machine learning models~\cite{pagano2023bias, mehrabi2021survey, chouldechova2017fair}. Verma~\etal~\cite{verma2018fairness} introduced three categories of fairness parameters: statistical measures, similarity-based measures, and causal reasoning. Additionally, Mittal~\etal~\cite{mittal2023bias} introduced metrics, such as parity-based metrics, score-based metrics, and facial-analysis-specific metrics. However, their study was not specialized for the FER task, and a thorough study of core models, such as CNNs and Transformer-based models, along with the state-of-the-art models in FER could open insight into bias mitigation.

Although some researchers introduced their own metrics, most studies have concentrated on widely recognized metrics such as Equal Odds (Eq-Od)~\cite{hardt2016equality}, Equal Opportunity (Eq-Op)~\cite{hardt2016equality}, Demographic Parity (De-Pa)~\cite{dwork2012fairness, kusner2017counterfactual}, and Treatment Equality (Tr-Eq)~\cite{berk2021fairness}. In this study, we focus on these four metrics to evaluate the bias and fairness of the FER models. To study bias in the models, we need to know the concept of protected and unprotected groups. Protected groups are privileged against discrimination, whereas unprotected groups may face exclusion or bias. In continuing, we quickly review the four bias metrics we will utilize in this paper. Table~\ref{tbl:model_metrics_summary} summarizes the metrics we use for the model analysis.
\\
\textbf{Equalized Odds (Eq-Od)}~\cite{hardt2016equality} evaluates fairness by examining the positive predicted samples. Eq.~\ref{eq:equalized_odds} states that for an attribute $A$, a fair model will produce equal ratios of True Positives (TP) and False Positives (FP) for both the protected group ($A\mathord{=}a$) and the unprotected group ($A\mathord{=}b$):
\begin{equation}
\label{eq:equalized_odds}
\begin{adjustbox}{width=0.80\columnwidth}
$\begin{aligned}
p(\hat{Y}=1|A=a, Y=y) = p(\hat{Y}=1|A=b, Y=y).
\end{aligned}$
\end{adjustbox} 
\end{equation}
\textbf{Equal Opportunity (Eq-Op)}~\cite{hardt2016equality} focuses exclusively on TPs, where the probability of TPs ($Y\mathord{=}1$) for protected and an unprotected groups should be the same. The key difference between this metric and Equalized Odds is illustrated in Eq.~\ref{eq:equal_opportunity}. While Equalized Odds consider both TPs and FPs, Equal Opportunity concentrates on TPs:
\begin{equation}
\label{eq:equal_opportunity}
\begin{adjustbox}{width=0.80\columnwidth}
$\begin{aligned}
p(\hat{Y}=1|A=a, Y=1) = p(\hat{Y}=1|A=b, Y=1).
\end{aligned}$
\end{adjustbox} 
\end{equation}
\textbf{Demographic Parity (De-Pa)}~\cite{dwork2012fairness, kusner2017counterfactual} examines the independence of predictions from attributes. In essence, it evaluates the distribution of positive predictions, irrespective of whether they are TP or FP. As Eq.~\ref{eq:demographic_parity} shows, this metric expects a uniform distribution of predictions across different groups:
\begin{equation}
\label{eq:demographic_parity}
\begin{adjustbox}{width=0.60\columnwidth}
$\begin{aligned}
p(\hat{Y}=1|A=a) = p(\hat{Y}=1|A=b).
\end{aligned}$
\end{adjustbox} 
\end{equation}
\textbf{Treatment Equality (Tr-Eq)}~\cite{berk2021fairness} examines the distribution of errors across different demographic groups (see Eq.~\ref{eq:treatment_equality}). Specifically, it ensures that the ratio of False Negatives (FNs) to False Positives (FPs) remains consistent between protected and unprotected groups:
\begin{equation}
\label{eq:treatment_equality}
\begin{adjustbox}{width=0.80\columnwidth}
$\begin{aligned}
\frac{p({\hat{Y}}=0|A=a, Y=1)}{p({\hat{Y}}=1|A=a, Y=0)} = \frac{p({\hat{Y}}=0|A=b, Y=1)}{p({\hat{Y}}=1|A=b, Y=0)}.
\end{aligned}$
\end{adjustbox} 
\end{equation}

To obtain the bias score for each model, we first applied all four bias metrics to each expression label separately, treating the target expression as the positive class and grouping all remaining expressions as the negative class. Next, we selected the maximum bias across all expressions as the bias score for each metric (see Eq.~\ref{eq:bias_score_metric}). To assess the bias of each demographic attribute, we then averaged these metric-specific bias scores using Eq.~\ref{eq:bias_score_attribute}. Finally, the overall bias of each model was computed by averaging the bias scores across all attributes (Eq.~\ref{eq:bias_score}).
\begin{equation}
\label{eq:bias_score_metric}
\begin{adjustbox}{width=0.76\columnwidth}
$\begin{aligned}
Bias_{att}(met) = Max(Bias_{met, att}(y))~~~~\forall~y\in Y,
\end{aligned}$
\end{adjustbox} 
\end{equation}
\begin{equation}
\label{eq:bias_score_attribute}
\begin{adjustbox}{width=0.78\columnwidth}
$\begin{aligned}
Bias(att) = \frac{1}{m}\sum Bias_{att}(met)~~~~\forall~met \in MET,
\end{aligned}$
\end{adjustbox} 
\end{equation}
\begin{equation}
\label{eq:bias_score}
\begin{adjustbox}{width=0.61\columnwidth}
$\begin{aligned}
Bias = \frac{1}{d}\sum Bias(att)~~~~\forall~att \in ATT,
\end{aligned}$
\end{adjustbox} 
\end{equation}
where $m$ indicates number of metrics and $d$ determines number of attributes ($m\mathord{=}4$ and $d\mathord{=}3$, in our study). To sum up, we employed four well-established bias metrics to evaluate fairness of seven FER models over three demographic attributes, age, gender, and race. The detailed discussion of the models' bias will follow in Section~\ref{sec:experimental_results}.

It is notable that FER is inherently a multi-class problem, while the fairness measures used in this research are defined for binary classification. To handle this constraint, we adopt a one-vs-rest formulation in which each expression is treated as the positive class and all remaining expressions are grouped as the negative class. This approach allows each fairness metric to be computed consistently across classes while preserving their probabilistic interpretation. We select the maximum bias across classes to highlight the most disadvantaged expression group, as mean aggregation may obscure severe class-specific disparities. 


\section{Experimental Results} 
\label{sec:experimental_results}
This section explores bias across four well-known FER datasets: AffectNet~\cite{mollahosseini2017affectnet}, Fer2013~\cite{goodfellow2013challenges}, RAF-DB~\cite{li2017reliable}, and ExpW~\cite{zhang2018facial}. Additionally, it examines fairness in seven deep models, including three CNN models: MobileNet~\cite{howard2017mobilenets}, ResNet~\cite{he2016deep}, XceptionNet~\cite{chollet2017xception}, as well as two Transformer-based models: ViT~\cite{dosovitskiy2020image} and CLIP~\cite{radford2021learning}, plus two state-of-the-art FER-specialized models: POSTER~\cite{zheng2023poster} and CEPrompt~\cite{zhou2024ceprompt}.

\subsection{Bias in Datasets} 
\label{sec:experimental_results:data_experiments}
Many studies suggest that the primary sources of bias in machine learning models lie in the data itself~\cite{barocas2023fairness, mehrabi2021survey, suresh2019framework}. Consequently, to investigate bias in FER, it is essential to first examine the datasets. Numerous facial attributes can contribute to bias~\cite{terhorst2021comprehensive}, and a variety of biases can be present within the datasets~\cite{cheong2023causal, mehrabi2021survey}. Our study examines not only the three primary demographic attributes (age, gender, and race) but also explores less visible sources of bias in FER datasets. These include variations in illumination, background noise, camera resolution, gestures, eye gaze, head pose, hair color, and other latent factors that may inadvertently amplify bias in the datasets and downstream models. In the following sections, we provide an overview of the datasets used in our study and detail the preprocessing steps we applied. Additionally, we explore the data distribution, attributes co-occurrence, and model-based analysis.

\subsubsection{Datasets} 
\label{sec:experimental_results:data_experiments:datasets}
Although numerous FER datasets exist, only a few feature in-the-wild facial images. By considering factors such as popularity, number of expression labels, and annotation methods, we selected four datasets for our study: AffectNet~\cite{mollahosseini2017affectnet}, Fer2013~\cite{goodfellow2013challenges}, RAF-DB~\cite{li2017reliable}, and ExpW~\cite{zhang2018facial}. All these datasets include at least six Ekman~\cite{ekman1971constants} basic expression labels, including Happy, Sad, Surprise, Fear, Disgust, Anger, plus Neutral, manually annotated by annotators.

\begin{table*}[b!] 
\caption{Bias metrics for age, gender, and race across datasets (in percent) include the Wasserstein Distance (WD), Jensen–Shannon Divergence (JSD), Conditional-Entropy Bias Index (CEBI), Concentration Index (CI), Normalized Shannon Entropy (NSE), Normalized Label Skewness (NLS), and Group-Normalized Mutual Information (GNMI). The mean score represents the average across the three attributes, while the bias score denotes the average of all metrics.}
\label{tbl:dataset_bias_score}
\centering
\small
\setlength{\tabcolsep}{6pt} 
\resizebox{01.0\textwidth}{!}
{
\begin{tabular}{l>{\color{mediumgray}}c>{\color{mediumgray}}c>{\color{mediumgray}}c>{\color{mediumgray}}c|>{\color{mediumgray}}c>{\color{mediumgray}}c>{\color{mediumgray}}c>{\color{mediumgray}}c|>{\color{mediumgray}}c>{\color{mediumgray}}c>{\color{mediumgray}}c>{\color{mediumgray}}c|>{\color{mediumgray}}c>{\color{mediumgray}}c>{\color{mediumgray}}c>{\color{mediumgray}}c}
\toprule

\multirow{2}{*}{} & \multicolumn{4}{c}{\textbf{RAF-DB}} & \multicolumn{4}{c}{\textbf{Fer2013}} & \multicolumn{4}{c}{\textbf{ExpW}} & \multicolumn{4}{c}{\textbf{AffectNet}} \\ 
\midrule

          & \color{black}Age & \color{black}Gender & \color{black}Race & \color{black}Mean & \color{black}Age & \color{black}Gender & \color{black}Race  & \color{black}Mean & \color{black}Age & \color{black}Gender & \color{black}Race & \color{black}Mean & \color{black}Age & \color{black}Gender & \color{black}Race & \color{black}Mean\\ 
\midrule
WD   & 5.3  & 3.1  & 3.1  & 3.8  & 4.9  & 3.6  & 3.6  & 4.0  & 5.0  & 6.0  & 2.7  & 4.6  & 5.3  & 2.7  & 3.0  & 3.6  \\
JSD  & 2.2  & 0.4  & 1.0  & 1.2  & 2.2  & 0.5  & 1.3  & 1.3  & 2.2  & 1.1  & 0.7  & 1.3  & 2.2  & 0.4  & 0.9  & 1.2  \\
NSE  & 5.2  & 10.3 & 3.5  & 6.3  & 8.3  & 16.3 & 5.5  & 10.0 & 2.8  & 5.2  & 1.8  & 3.3  & 1.6  & 5.4  & 1.1  & 2.7  \\
NLS  & 54.0 & 51.4 & 51.8 & 52.4 & 30.0 & 49.9 & 32.2 & 37.4 & 47.6 & 50.1 & 46.8 & 48.2 & 58.5 & 54.2 & 51.3 & 54.7 \\
GNMI & 61.9 & 29.3 & 76.0 & 55.8 & 54.7 & 26.1 & 70.2 & 50.4 & 64.4 & 31.7 & 83.7 & 59.9 & 67.1 & 33.3 & 84.3 & 61.6 \\
CEBI & 82.2 & 63.7 & 90.5 & 78.8 & 83.1 & 64.8 & 89.4 & 79.1 & 82.0 & 61.5 & 91.0 & 78.2 & 81.6 & 62.5 & 90.3 & 78.2 \\
CI   & 13.0 & 9.5  & 14.8 & 12.4 & 14.1 & 11.2 & 15.0 & 13.4 & 12.6 & 7.0  & 14.8 & 11.5 & 11.9 & 7.4  & 14.2 & 11.2 \\
\color{black}Bias Score & \color{black}31.9 & \color{black}23.9 & \color{black}34.3 & \color{black}30.1 & \color{black}28.1 & \color{black}24.6 & \color{black}31.0 & \color{black}27.9 & \color{black}30.9 & \color{black}23.2 & \color{black}34.5 & \color{black}29.5 & \color{black}32.6 & \color{black}23.7 & \color{black}35.0 & \color{black}30.4 \\
\bottomrule
\end{tabular}
}
\end{table*}

AffectNet is the largest in-the-wild FER dataset, comprising nearly 1 million images, with half of them manually annotated. It includes eight expressions (six basic expressions, plus Contempt and Neutral), continuous labels (valence-arousal), and additional metadata, reflecting the broader duality between categorical and dimensional emotion representations explored in affective computing~\cite{hosseini2024deep}. Fer2013 contains approximately 36K images, divided into 29K training samples and 3.5K validation samples. However, its small image size (40$\times$40 pixels) and grayscale color representation pose limitations. RAF-DB provides both single-label and compound-label subsets, with around 30K facial images and facial landmark annotations. The compound-labeling approach in RAF-DB results in 19 expressions, including the seven basic expressions and 12 compound expressions such as Happily-Surprise and Sadly-Angry. Finally, ExpW is a large-scale FER dataset, featuring over 90K manually annotated facial images. We utilize these datasets in our study to train and evaluate our models.

\begin{figure*}[t!]
\centering
\includegraphics[width=1.0\textwidth]{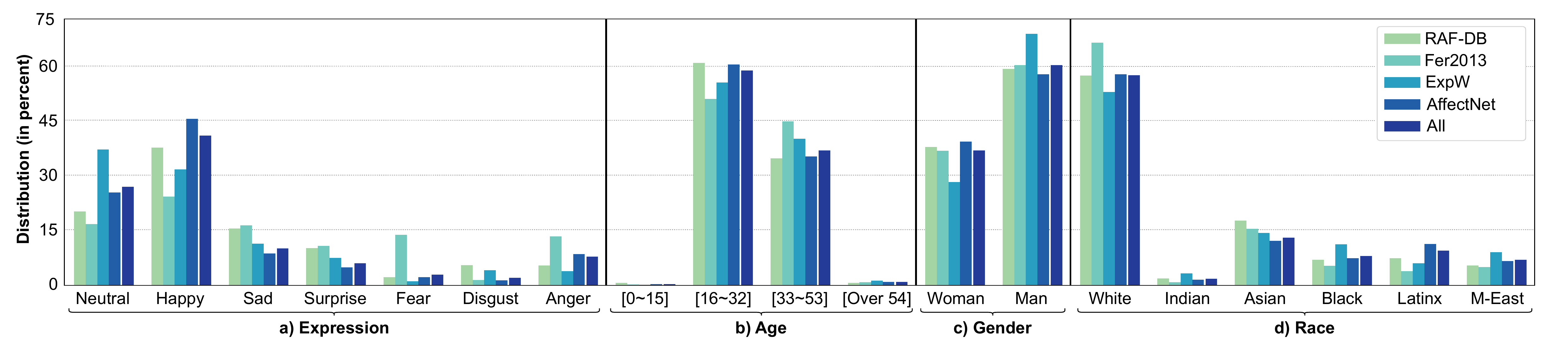}
\caption{The data distribution across different datasets shows several trends: a) Happy and Neutral dominate the datasets, while Fear and Disgust are underrepresented. Among all datasets, Fer2013 exhibits the most balanced expression distribution, b) A noticeable bias is observed in the age groups, where [16$\sim$32] and [33$\sim$53] being more frequent, while [0$\sim$15] and [Over 54] have significantly fewer samples, c) Across all the datasets, there are more Man samples than Womans. This gender imbalance is most pronounced in the ExpW dataset and least evident in AffectNet, d) Regarding race, White samples are the most represented group, while Indians are the least represented. Data distribution for Black, Latinx, and Middle-Eastern races is more even.}
\label{fig:data_distribution}
\end{figure*}

\subsubsection{Data Preprocessing} \label{sec:experimental_results:data_experiments:data_preprocessing}
To have a unique form of data, in the first step, we aligned images by ensuring the same cropping area, image size, and color domain across all the datasets. We extended the original crop region by 25 to 35 percent in each dimension for all images, then used the DeepFace model~\cite{serengil2024lightface, serengil2020lightface} to process the extended crop and align the facial images. Next, we resized all the cropped faces to $224\times224$. Since Fer2013 is grayscale, we replicated the grayscale layer three times to match the RGB format, ensuring uniformity across the data. Additionally, ExpW lacks a validation set, therefore, we randomly selected 10 percent of its images to serve as the validation set.

After that, we used the DeepFace~\cite{serengil2021hyperextended} model to extract facial attributes, including age, gender, and race. Ages were categorized into four ranges: [0$\sim$15], [16$\sim$32], [33$\sim$53], and [Over 54]. DeepFace predicts six racial categories: White, Black, Asian, Indian, Latinx, and Middle Eastern. For gender, the model outputs either Woman or Man. Throughout the remainder of this paper, we will use these terms to refer to the groups.

It is notable that DeepFace is used solely to assign demographic categories for bias analysis and is never involved in training or optimizing any FER model, ensuring that its predictions do not influence model behavior. Although automated demographic estimators may contain biases, applying the same estimator uniformly across all datasets and models maintains the validity of our comparative evaluations, since any estimator-specific inaccuracies affect all groups equally rather than altering relative differences across datasets or architectures.

\subsubsection{Metric-Based Dataset Analysis}
\label{sec:experimental_results:data_experiments:metric_based_dataset_analysis}
In the initial experiment, we assessed dataset bias using the customized metrics described in Sec.~\ref{sec:methodology:data_analysis}, including Wasserstein Distance (WD), Jensen–Shannon Divergence (JSD), Normalized Shannon Entropy (NSE), Normalized Label Skewness (NLS), Group Normalized Mutual Information (GNMI), as  well as two new metrics of Conditional-Entropy Bias Index (CEBI) and Concentration Index (CI). By applying Eqs.~\ref{eq:wasserstein_distance} to~\ref{eq:concentration_index}, we derived the bias scores summarized in Table~\ref{tbl:dataset_bias_score}. This table presents the scores for each metric and demographic attribute, where the final bias score of each dataset is calculated over the average of different scores.

The bias analysis of the four datasets reveals notable demographic disparities, with overall bias scores ranging from 27.9\% (Fer2013) to 30.4\% (AffectNet). None of the datasets are fully balanced across age, gender, and race, suggesting that models trained on these datasets are likely to inherit these imbalances. Among the attributes, race consistently exhibits the highest bias values, indicating that racial representation is more uneven than age and gender. Gender shows comparatively lower bias, although metrics like CEBI and NLS indicate persistent disparities, while age-related bias is moderate but consistent. These findings highlight that demographic representation is far from uniform, and differences in attribute distribution could affect fairness in model predictions.

Examining individual metrics provides deeper insight into the nature of these biases. Metrics such as CEBI and GNMI report high values across all datasets, indicating strong conditional dependencies and unequal mutual information between demographic attributes and labels. CI values reveal the uneven distribution of samples across different attribute-label combinations, while NSE and NLS reflect deviations from uniform label distributions, with Fer2013 showing higher unevenness in age and gender, and AffectNet showing pronounced skew in age. WD and JSD are relatively lower, suggesting that while overall distributional differences exist, they are less pronounced than the conditional dependencies captured by CEBI and GNMI. Overall, these results underscore the importance of considering both dataset composition and metric-specific insights when addressing bias in facial expression recognition.

\begin{figure*}[t]
\centering
\includegraphics[width=0.92\textwidth]{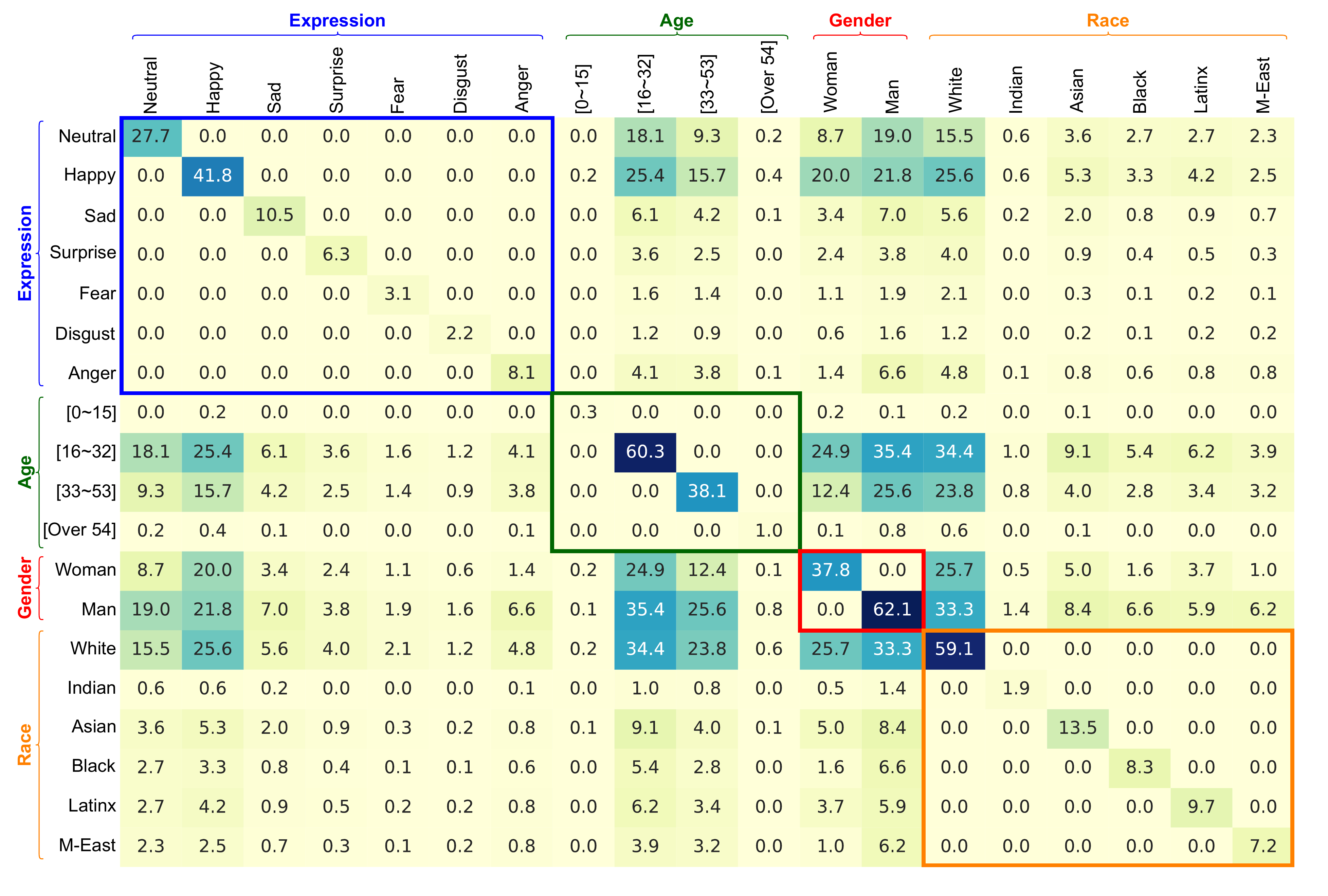}
\caption{This 2D co-occurrence matrix illustrates the relationships between different attributes across all the datasets (in percent). The diagram visualizes the data distribution for each attribute value and highlights the contribution of each attribute to others. This heat map reveals biases toward Neutral and Happy. The age groups [16$\sim$32] and [33$\sim$53] are the most prominent, while Man appears the most frequent gender. In terms of race, White is overrepresented, whereas Indian is underrepresented.}
\label{fig:correlation_matrix_2D}
\end{figure*}

\subsubsection{Attributes Distribution}
\label{sec:experimental_results:data_experiments:attribures_distribution}
In a subsequent experiment, we also studied Macro-Level Co-occurrence Bias by analyzing probability distribution of data. Fig.~\ref{fig:data_distribution}-a highlights that Neutral and Happy are the most prevalent expressions, whereas Fear and Disgust are the least represented. Notably, Fer2013 and RAF-DB demonstrate a more balanced distribution of expressions compared to the pronounced imbalances observed in AffectNet and ExpW. 
Fig.~\ref{fig:data_distribution}-b reveals a pronounced bias in dataset across age groups, where age ranges of [16$\sim$32] and [33$\sim$53] are significantly overrepresented. In contrast, samples from [0$\sim$15] and [Over 54] age groups account for less than 5\% of the total data in all datasets.

As shown in Fig.~\ref{fig:data_distribution}-c, the number of Man samples is approximately 50\% more than Woman, though this gender bias is less pronounced compared to other attributes. Fig.~\ref{fig:data_distribution}-d, on the other hand, highlights a notable racial imbalance, with White samples constituting over 50\% of the total across all datasets. The distribution of Asian, Black, Latinx, and Middle Eastern samples is more balanced, while Indian samples remain consistently underrepresented. In conclusion, an examination of the data distribution across four essential attributes of age, gender, race, and expression, uncovers notable bias in the datasets, where addressing them  is vital to ensure fairness in FER models.

\subsubsection{Attributes Co-occurrence}
\label{sec:experimental_results:data_experiments:attribures_correlation}
Building on Micro-Level Co-occurrence Bias, we studied the joint probability distribution of data. Examining relations between joint attributes provides a meaningful approach for data analysis. We kept expression as the primary attribute and analyzed its occurrence with other facial attributes: age, gender, and race.

Figure~\ref{fig:correlation_matrix_2D} illustrates that 27.7\% of the data is labeled Neutral, where 18.1\% belong to the [16$\sim$32] age group, and 9.3\% fall within [33$\sim$53]. In contrast, less than 0.3\% of Neutral label corresponds to the two remaining age groups. Regarding gender distribution, Neutral reveals a significant disparity between Man and Woman samples, with the number of Man being more than double that of Woman. For racial distribution within this expression, 15.5\% of the samples are associated with White, while the total representation of the other five racial groups is below 13\%.

The distribution of Happy exhibits a similar bias toward the age groups [16$\sim$32] and [33$\sim$53]. However, no significant gender disparity is observed between Man and Woman. In contrast, bias exists toward White group for this expression. Compared to Happy and Neutral, the bias across attributes is less pronounced for the other expressions, particularly across age groups and racial categories. Among the datasets, ExpW demonstrates the highest bias, while Fer2013 emerges as the least biased.

\begin{figure*}[t]
\centering
\includegraphics[width=1.0\textwidth]{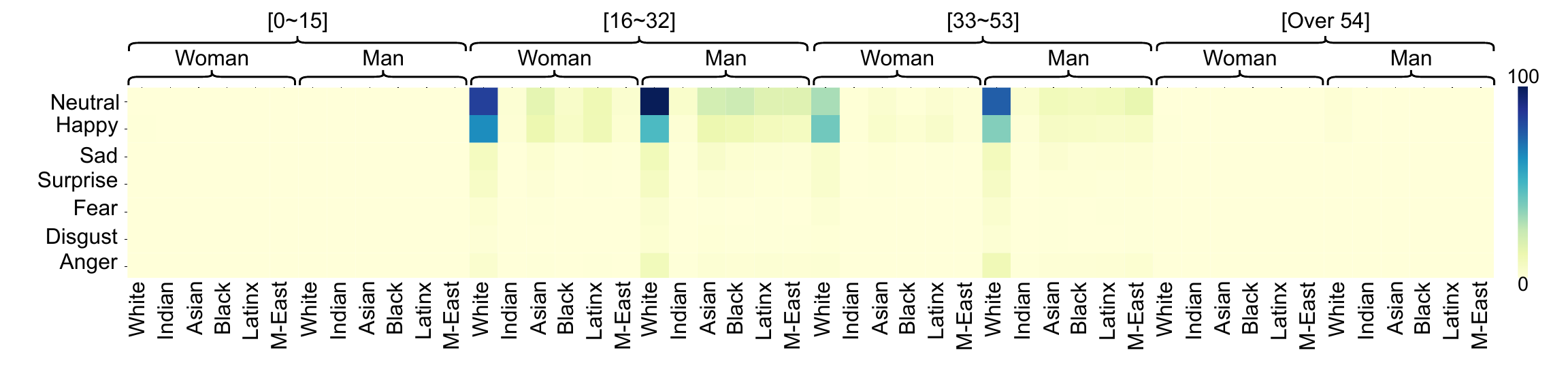}
\caption{The 4D co-occurrence matrix between different attributes of all the datasets is presented, with expressions represented in the rows and the columns divided first by age groups, followed by gender, and finally by race groups. This heat map highlights an uneven data distribution, where a great portion of data are underrepresented. This diagram reveals limited and imbalance diversity in the datasets.}
\label{fig:correlation_matrix_4D}
\end{figure*}

Full joint attribute co-occurrence, Fig.~\ref{fig:correlation_matrix_4D}, offers deeper insights into bias in datasets.  This figure reveals that a great portion of data are concentrated on only 8 out of 336 possible combination of attributes. This figure highlights high dataset bias through Happy and Neutral expressions, [16$\sim$32] and [33$\sim$53] age groups, Man gender, and White race. Happy and Neutral account for 60.7\% of the data distribution, 98.6\% of the samples fall within the [16$\sim$53] age ranges, 62.6\% of the samples are labeled as Man, and White dominates with 59.3\% of the racial data distribution. 

\subsubsection{Dataset Generality}
\label{sec:experimental_results:data_experiments:dataset_generality}
Based on the data distribution and co-occurrence analysis, we examined demographic biases within the datasets. While demographic disparity is a primary source of bias, other factors such as illumination, head pose, image quality, background settings, lighting effects, and facial accessories also contribute to bias in dataset~\cite{wang2026occlusion, srinivas2019face, singh2022anatomizing, udefi2023analysis}. These sources of bias vary significantly across different datasets.

\begin{figure}[b!]
\centering
\includegraphics[width=0.50\columnwidth]{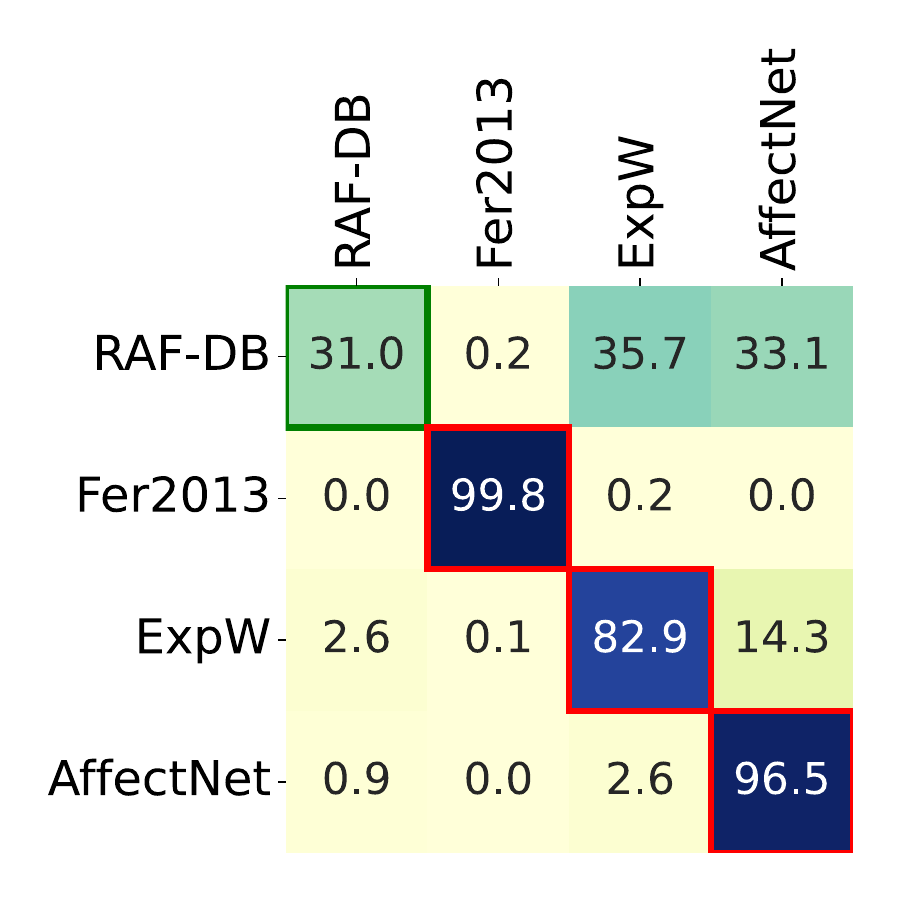}
\caption{The confusion matrix for FER datasets (in percent) reveals that the high diagonal values for Fer2013, ExpW, and AffectNet point to inherent biases within these datasets. Notably, RAF-DB stands out as the only dataset that shows a significant correlation with the others.}
\label{fig:confusion_matrix_FER_datasets}
\end{figure}

To investigate general bias in FER datasets, we trained an XceptionNet model~\cite{chollet2017xception} to identify the dataset origin of each image by assigning dataset labels as training targets. The results, as shown in Fig.~\ref{fig:confusion_matrix_FER_datasets}, demonstrated biases inherent to FER datasets. While some correlation exists between RAF-DB and two other datasets (ExpW and AffectNet), Fer2013, ExpW, and AffectNet remain uncorrelated, indicating biases inherent to these datasets. This high dataset-classification accuracy reveals the presence of “dataset-signature” biases, which relates to differences in resolution, cropping ratios, background texture, lighting patterns, compression artifacts, and collection protocols. These dataset-specific signatures allow a model to identify the source dataset even when demographic attributes are balanced, confirming that FER datasets differ significantly in their visual characteristics. Such implicit dataset cues can propagate into FER model predictions, highlighting a relatively underexplored yet important source of bias in FER research.

An additional experiment was conducted to investigate the generality and biases in the datasets. This generalization gap mirrors challenges reported in cross-dataset FER domain-adaptation work, where distribution shift between source and target datasets substantially degrades transfer accuracy unless explicitly modeled~\cite{yan2026domain}. In this setup, one dataset was excluded, and a facial expression recognition model (XceptionNet) was trained on the remaining datasets. For fairness, the model was first evaluated on the validation sets of the included datasets, then tested on the excluded dataset to assess its generalizability. Since the excluded dataset was not used in training, the results reflect its bias. Table~\ref{tbl:leave_one_dataset_out_accuracy} compares validation accuracies on included datasets with test accuracy on the excluded one, to study the bias of the datasets used for training. Interestingly, test accuracies for RAF-DB and Fer2013 were higher than their validation accuracies, indicating that ExpW and AffectNet used in training improved generalization. The opposite trend was observed for ExpW and AffectNet, suggesting that training on RAF-DB and Fer2013 failed to capture their challenges. For example, when AffectNet was excluded, validation accuracy across RAF-DB, Fer2013, and ExpW was 57.0\%, while test accuracy on AffectNet dropped to 45.9\%. Overall, this experiment suggests higher inherent bias in RAF-DB and Fer2013 and lower bias in ExpW and AffectNet.

It is important to note that the leave-one-dataset-out results reflect not only dataset-level generalization but also the inherent difficulty of predicting different expressions. Datasets with a higher proportion of easy expressions (e.g., Happy) may show inflated generalization, whereas those with noisier or harder expressions (e.g., Fear, Disgust) may yield lower accuracy. Thus, part of the performance gap may stem from expression difficulty rather than dataset bias alone. Considering both factors provides a more accurate view of the generality and limitations of FER datasets.

\begin{table}[b] 
\caption{Bias analysis across different datasets (in \%). When one dataset is excluded, a model is trained on the remaining datasets. The model is evaluated on the validation set of the included datasets, and then is tested over the excluded dataset. More difference between the validation and test scores highlights more inherent biases in the datasets used for the training.}
\label{tbl:leave_one_dataset_out_accuracy}
\centering
\small
\setlength{\tabcolsep}{12pt} 
\resizebox{1.0\linewidth}{!}
{{
\begin{tabular}{lccccc}
\toprule
            & \multicolumn{4}{c}{\textbf{Excluded on Training}}\\
            \cmidrule{2-5}
            & \textbf{RAF-DB} & \textbf{Fer2013} & \textbf{Exp}W & \textbf{AffectNet} \\ \midrule
Validation  & 59.0 & 54.1 & 59.0 & 57.0\\
Test        & 60.1 & 71.2 & 44.0 & 45.9 \\
\bottomrule
\end{tabular}
}}
\end{table}

\subsubsection{Conclusion}
\label{sec:experimental_results:data_experiments:conclusion}
Our findings reveal several key factors driving dataset bias. First, demographic imbalance is the primary source, particularly for age and race. Second, conditional dependencies between emotions and demographics, captured by CEBI and GNMI, show that some expressions are overrepresented within specific groups. Third, intra-group imbalance, quantified by CI and NLS, indicates uneven label distributions even within a single demographic. Finally, model-based experiments confirm that dataset-specific characteristics such as lighting, background, gestures, and cultural differences in emotional interpretation introduce non-demographic bias and hinder generalization. Overall, these results highlight the multifaceted nature of FER dataset bias and motivate more balanced, context-aware data collection strategies.

\subsection{Fairness of Models} 
\label{sec:experimental_results:model_experiments}
After analyzing bias in the FER datasets, we further investigate bias and fairness within the models. As mentioned, four bias metrics are used to evaluate seven deep models with different architectures. In the following sections, we describe our training methodology and discuss the accuracy and bias of each model.

\subsubsection{Training and Implementation Details}
We selected five CNN- and Transformer-based models (MobileNet, ResNet, XceptionNet, ViT, and CLIP) and two SOTA FER-specialized models (POSTER and CEPrompt). MobileNet served as a baseline CNN due to its streamlined feed-forward structure without residual connections, making it computationally efficient for mobile and embedded use. ResNet and XceptionNet, with residual connections, were chosen for their ability to mitigate vanishing gradients and improve accuracy. Among Transformers, ViT represents a state-of-the-art vision approach using self-attention instead of convolution, while CLIP, a large vision-language model, enables joint image–text processing. To further analyze FER bias, we evaluated POSTER, which combines self-attention with posterior modeling to capture subtle emotions, and CEPrompt, which uses context-aware prompts to guide Vision Transformers for more robust recognition.

\begin{figure*}[t]
\centering
\includegraphics[width=1.0\textwidth]{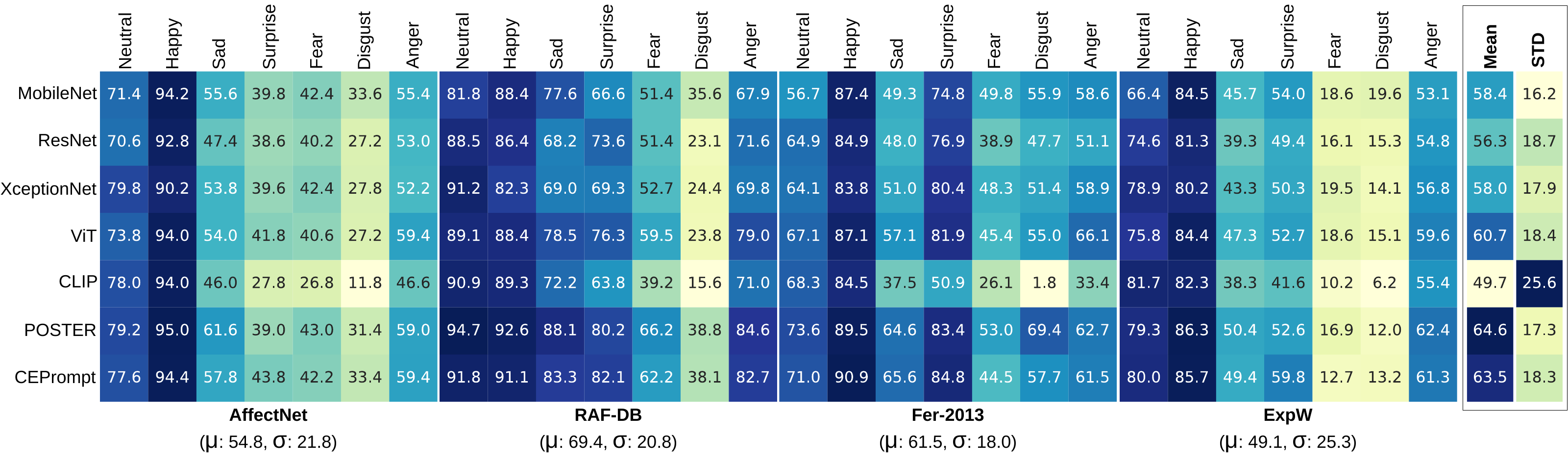}
\caption{Model accuracy for each expression (in percent). In this experiment, models were trained on individual datasets and evaluated on their respective evaluation sets. $\mu$ and $\sigma$ denote the average accuracy and its standard deviation for each dataset, while Mean and STD represent the overall average accuracy and standard deviation across each model. RAF-DB emerged as the least challenging dataset, whereas ExpW posed the most difficulty. Among the models, POSTER reached the highest accuracy, contrasting with CLIP which showed the lowest accuracy.}
\label{fig:accuracy_matrix_of_models_over_datasets}
\end{figure*}

To support reproducibility, this experiment was conducted using Python 3.8.10, TensorFlow 2.9.2~\cite{abadi2016tensorflow}, and the Hugging Face~\cite{wolf2020transformers} libraries. Four Nvidia GPUs with 8 GB and 12 GB of memory were utilized. As a preprocessing step, image dimensions were fixed at $224\times224$. We conducted experiments on each dataset and model separately and also merged the training and validation sets of all datasets for subsequent experiments. Notably, after merging the datasets, we obtained 404,755 training samples and 22,858 validation samples, while the minimum of one thousand samples was considered for each group, to make sure about the richness of the groups. 

In the first experiment, we evaluated the accuracy of each model on every dataset, separately. In the second, we merged all training sets, trained each model, and then measured accuracy on the combined validation set. Finally, the third experiment focused on assessing each model's bias utilizing the bias metrics.  For this, we fully trained four models, including MobileNet, ResNet, XceptionNet, and ViT, and fine-tuned CLIP, POSTER, and CEPrompt. Full training was conducted for 30 epochs, while fine-tuning was limited to 5 epochs. We used the Adam optimizer with a learning rate of $10^{-3}$ for MobileNet, ResNet, XceptionNet, CLIP, and CEPrompt, and $10^{-5}$ for ViT and POSTER. Categorical cross-entropy served as the loss function for all training procedures.

\subsubsection{Facial Expression Recognition}
\label{sec:experimental_results:model_experiments:facial_expression_recognition}
Figure~\ref{fig:accuracy_matrix_of_models_over_datasets} reports model accuracy across datasets by expression. All models show a clear tendency toward Neutral and Happy, which consistently achieve higher accuracy than other expressions. Comparing the accuracy patterns with the expression distribution in Figure~\ref{fig:correlation_matrix_2D} indicates that data imbalance still contributes to this behavior. Notably, Neutral and Happy are typically easier to recognize, occur more frequently, and exhibit lower perceptual ambiguity, which naturally leads to higher model accuracy. This is consistent with prior FER findings that high-frequency, low-ambiguity expressions yield more stable feature representations~\cite{guo2018dominant}. Overall, the trend reflects a combination of expression difficulty, dataset imbalance, and model tendencies. Nonetheless, models partially compensated for imbalance; for example, although the Disgust-to-Happy sample ratio was highly skewed ($\frac{2.2}{41.8}\approx0.05$), the accuracy gap was much smaller ($\frac{29.1}{85.9}\approx0.33$), suggesting some mitigation during training.

\begin{figure*}[t]
\centering
\includegraphics[width=1.0\textwidth]
{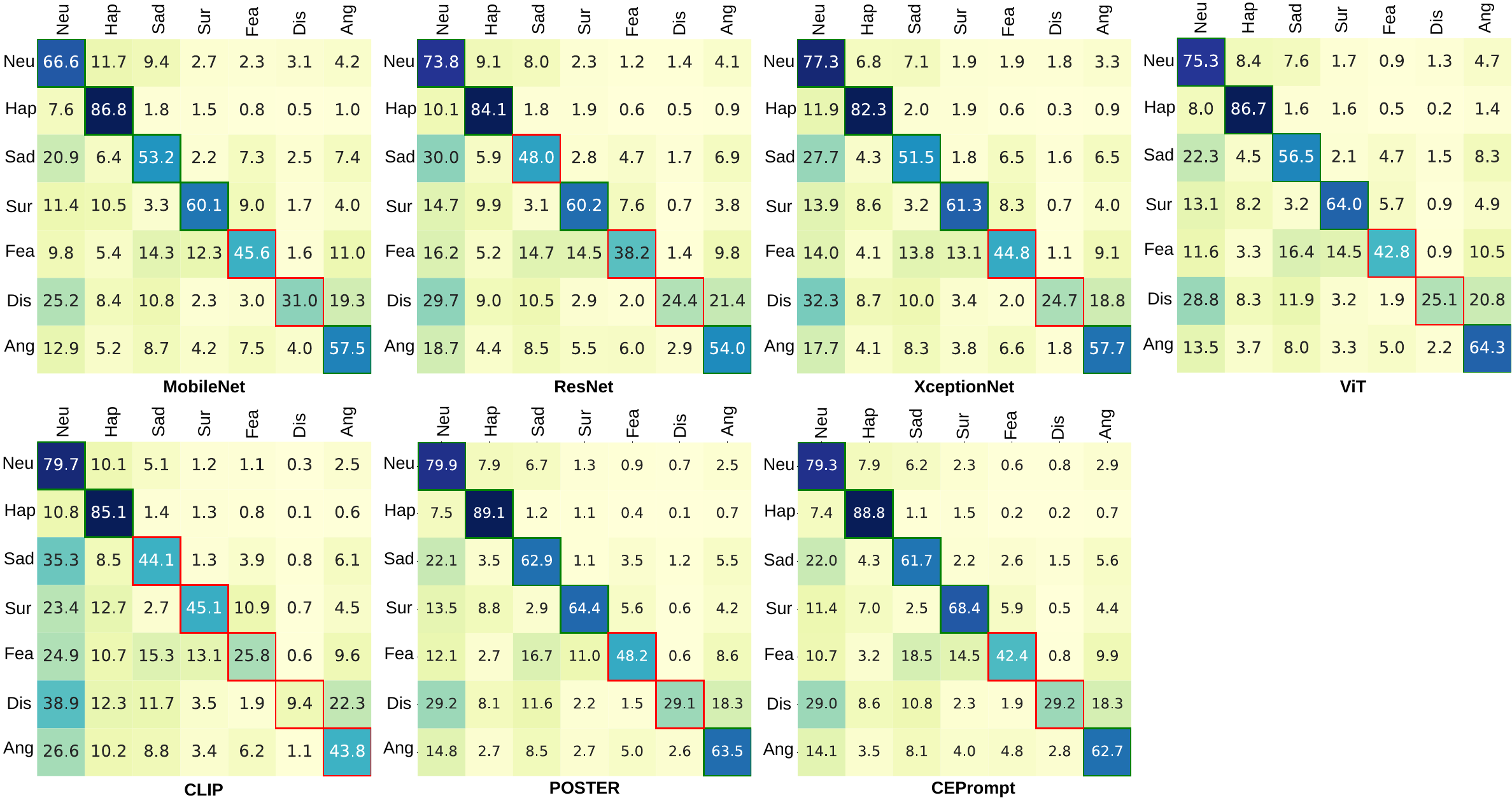}
\caption{Confusion matix of each model across all datasets (in percent). In this experiment, all the 
datasets are used for both of the training and validation stages. No model was able to achieve high scores for the two challenging expressions, Fear and Disgust. A comparison of the results reveals that ViT demonstrated the most consistent accuracy score, while CLIP struggled to achieve satisfactory performance.}
\label{fig:confusion_matrix_FER_models}
\end{figure*}

The average accuracy for each dataset, represented by $\mu$ in Fig.~\ref{fig:accuracy_matrix_of_models_over_datasets}, demonstrates a correlation between dataset richness and complexity. Richer datasets tend to be more challenging, leading to lower average accuracies. The last two columns of this figure, labeled Mean and STD, summarize the mean accuracy and standard deviation for each model. Through this figure, POSTER and CEPrompt (two FER-specialized models) achieved the highest accuracy, which looks normal. Among the rest of five basic models, ViT achieved the highest accuracy. While this suggests POSTER, CEPrompt and ViT may be the fairest models, subsequent experiments reveal they exhibit significant bias.

In contrast, the CLIP model showed the worst performance, with an average accuracy of 49.7\% and a high standard deviation exceeding 25\%. These results indicate that the CLIP model is not well-optimized for FER tasks in its current form. Training the entire model, rather than just fine-tuning its decision-making head, could potentially improve its performance. Lastly, among the fully trained models (MobileNet, ResNet, XceptionNet, and ViT), ViT achieved the highest accuracy score, while ResNet exhibited the lowest standard deviation. Based solely on high accuracy and low standard deviation metrics, the models can be ranked from least to most accurate models as follows: POSTER, CEPrompt, ViT, ResNet, MobileNet, XceptionNet, and CLIP. We will discuss this ranking in Sec.~\ref{sec:experimental_results:model_experiments:conclusion} when we extract the metric-based bias scores.

Another experiment for FER models involved training each model using the training samples from all datasets and evaluating their performance on all validation sets. Fig.~\ref{fig:confusion_matrix_FER_models} displays the confusion matrix for these general models across all validation sets. At first glance, it is clear that Fear and Disgust are the most challenging expressions for all models, with none achieving more than 50\% accuracy in predicting them.

Deeper analysis of Fig.~\ref{fig:confusion_matrix_FER_models} reveals significant confusion among certain expressions. For instance, pairs like Sad-Neutral, Disgust-Neutral, and Disgust-Anger pose considerable challenges, with erroneous predictions in the range of $18.8\%-35.3\%$, while zero values would be expected. The expressions accuracy below 50\% are highlighted by red color in this figure. A comparison between POSTER and CEPrompt models and the rest of the models illustrates that they could not tackle the problem of Fear and Disgust expressions in FER task. The general model results, summarized in Table~\ref{tbl:general_accuracy_fer_models}, indicate that the ViT model achieved the highest average accuracy in this experiment. Additionally, MobileNet had the lowest standard deviation, indicating more consistency in its predictions. While these experiments give us the clues of more fairness in the Transformer-based models, their results will be discussed in Sec.~\ref{sec:experimental_results:model_experiments:conclusion}.

\begin{table}[b!]
\caption{General accuracy of each model (in percent): four models MobileNet, ResNet, XceptionNet, and ViT are fully trained, while three models CLIP, POSTER, and CEPrompt are fine-tuned. In this experiment, the models are trained and evaluated over the combination of all the datasets.}
\label{tbl:general_accuracy_fer_models}
\centering
\small
\setlength{\tabcolsep}{4pt} 
\resizebox{0.99\linewidth}{!}
{{
\begin{tabular}{l>{\color{mediumgray}}c>{\color{mediumgray}}c>{\color{mediumgray}}c>{\color{mediumgray}}c>{\color{mediumgray}}c>{\color{mediumgray}}c>{\color{mediumgray}}cc}
\toprule
            & \color{black}\textbf{Neu} & \color{black}\textbf{Hap} & \color{black}\textbf{Sad}  \color{black}
            & \color{black}\textbf{Sur} & \color{black}\textbf{Fea} & \color{black}\textbf{Dis} & \color{black}\textbf{Ang}
            & \textbf{Mean ± STD}  \\
            \midrule
MobileNet   & 66.6 & 86.8 & 53.2 & 60.1 & 45.6 & 31.0 & 57.5  & ~57.3 ± 17.4 \\
ResNet      & 73.8 & 84.1 & 48.0 & 60.2 & 38.2 & 24.4 & 54.0  & ~54.7 ± 20.4 \\
XceptionNet & 77.3 & 82.3 & 51.5 & 61.3 & 44.8 & 24.7 & 57.7  & ~57.1 ± 19.6 \\
ViT         & 75.3 & 86.7 & 56.5 & 64.0 & 42.8 & 25.1 & 64.3  & ~59.2 ± 20.4 \\
CLIP        & 79.7 & 85.1 & 44.1 & 45.1 & 25.8 & 9.4  & 43.8  & ~47.5 ± 25.8 \\
POSTER      & 79.9 & 89.1 & 62.9 & 64.4 & 48.2 & 29.1 & 63.5  & \textbf{62.4 ± 18.2} \\
CEPrompt    & 79.3 & 88.8 & 61.7 & 68.4 & 42.4 & 29.2 & 62.7  & ~61.7 ± 18.9 \\
\bottomrule
\end{tabular}
}}
\end{table}

\begin{table}[t!] 
\caption{Bias analysis using the Equalized Odds metric, across models and three demographic attributes: age, gender, and race. Bias is calculated for each expression, followed by the calculation of maximum, mean, and standard deviation for each attribute. Range is [0, 100].}
\label{tbl:bias_equalized_odds}
\centering
\small
\setlength{\tabcolsep}{3pt} 
\resizebox{0.489\textwidth}{!}
{
\begin{tabular}{ll>{\color{mediumgray}}c>{\color{mediumgray}}c>{\color{mediumgray}}c>{\color{mediumgray}}c>{\color{mediumgray}}c>{\color{mediumgray}}c>{\color{mediumgray}}ccc}
\toprule
                                                   &        & \color{black}\textbf{Neu} & \color{black}\textbf{Hap} & \color{black}\textbf{Sad} 
                                                            & \color{black}\textbf{Sur} & \color{black}\textbf{Fea} & \color{black}\textbf{Dis} 
                                                            & \color{black}\textbf{Ang} & \textbf{Max} & \textbf{Mean ± STD} \\
\midrule
\multicolumn{1}{l}{}                              & Age    & 4.9 & 2.1 & 1.1 & 0.3 & 1.0 & 2.0 & 1.1 & 4.9 & ~1.7 ± 1.3 \\
\multicolumn{1}{l}{}                              & Gender & 7.7 & 12.7 & 2.2 & 2.3 & 0.1 & 5.9 & 2.8 & 12.7 & ~4.8 ± 3.9 \\
\multicolumn{1}{l}{\multirow{-3}{*}{MobileNet}}   & Race   & 11.6 & 9.0 & 7.1 & 5.9 & 2.2 & 7.7 & 4.7 & 11.6 & ~6.8 ± 2.8 \\ \midrule
\multicolumn{1}{l}{}                              & Age    & 6.7 & 1.8 & 0.7 & 0.4 & 1.7 & 0.6 & 2.1 & 6.7 & ~1.9 ± 2.0 \\
\multicolumn{1}{l}{}                              & Gender & 6.9 & 11.6 & 0.8 & 2.3 & 0.5 & 2.8 & 4.4 & 11.6 & ~4.1 ± 3.6 \\
\multicolumn{1}{l}{\multirow{-3}{*}{ResNet}}      & Race   & 10.1 & 6.3 & 7.1 & 4.7 & 1.4 & 5.0 & 3.4 & 10.1 & ~5.4 ± 2.5 \\ \midrule
\multicolumn{1}{l}{}                              & Age    & 5.1 & 2.0 & 0.9 & 0.5 & 1.9 & 1.0 & 1.1 & 5.1 & ~1.7 ± 1.4 \\
\multicolumn{1}{l}{}                              & Gender & 7.2 & 12.2 & 2.1 & 2.5 & 0.9 & 3.2 & 2.9 & 12.2 & ~4.4 ± 3.6 \\
\multicolumn{1}{l}{\multirow{-3}{*}{XceptionNet}} & Race   & 10.2 & 7.9 & 6.3 & 5.7 & 1.4 & 5.2 & 2.8 & 10.2 & ~5.6 ± 2.7 \\ \midrule
\multicolumn{1}{l}{}                              & Age    & 4.7 & 2.0 & 0.5 & 0.7 & 0.8 & 1.4 & 2.8 & 4.7 & ~1.8 ± 1.3 \\
\multicolumn{1}{l}{}                              & Gender & 9.5 & 13.4 & 0.5 & 2.9 & 0.4 & 2.8 & 4.8 & 13.4 & ~4.8 ± 4.4 \\
\multicolumn{1}{l}{\multirow{-3}{*}{ViT}}         & Race   & 13.1 & 9.6 & 9.1 & 6.0 & 1.5 & 4.3 & 5.2 & 13.1 & ~6.9 ± 3.5 \\ \midrule
\multicolumn{1}{l}{}                              & Age    & 7.3 & 2.8 & 0.1 & 0.9 & 2.7 & 0.1 & 2.6 & 7.3 & ~2.3 ± 2.3 \\
\multicolumn{1}{l}{}                              & Gender & 8.4 & 11.0 & 1.3 & 2.2 & 1.2 & 0.1 & 6.7 & 11.0 & ~4.4 ± 3.9 \\
\multicolumn{1}{l}{\multirow{-3}{*}{CLIP}}        & Race   & 12.5 & 7.4 & 7.4 & 4.9 & 4.7 & 1.1 & 5.6 & 12.5 & ~6.2 ± 3.2 \\ \midrule
\multicolumn{1}{l}{}                              & Age    & 6.4 & 1.1 & 0.4 & 0.5 & 2.1 & 0.4 & 2.8 & 6.4 & ~1.9 ± 2.0 \\
\multicolumn{1}{l}{}                              & Gender & 5.9 & 10.6 & 1.2 & 2.0 & 1.5 & 0.4 & 6.6 & 10.6 & ~4.0 ± 3.4 \\
\multicolumn{1}{l}{\multirow{-3}{*}{POSTER}}      & Race   & 10.8 & 6.9 & 6.9 & 4.8 & 5.6 & 2.1 & 6.3 & 10.8 & ~6.2 ± 2.4 \\ \midrule
\multicolumn{1}{l}{}                              & Age    & 5.1 & 1.0 & 0.1 & 0.6 & 2.0 & 0.3 & 3.0 & 5.1 & ~1.7 ± 1.6 \\
\multicolumn{1}{l}{}                              & Gender & 7.3 & 10.2 & 1.7 & 1.8 & 1.6 & 0.2 & 6.9 & 10.2 & ~4.2 ± 3.5 \\
\multicolumn{1}{l}{\multirow{-3}{*}{CEPrompt}}    & Race   & 12.4 & 6.3 & 6.1 & 5.4 & 4.4 & 2.1 & 6.4 & 12.4 & ~6.1 ± 2.9 \\
\bottomrule
\end{tabular}
}
\end{table}

So far, we have assessed the models accuracy to evaluate their robustness against bias stemming from imbalanced data distribution. Our findings indicate that, overall, POSTER and CEPrompt are the most accurate models, while CLIP performed poorly. Although these results suggest that POSTER, CEPrompt, and ViT might be the least biased models, subsequent experiments reveal different outcomes. In the next step, we will apply various bias metrics to further analyze the models biases across different attributes.

\subsubsection{Metric-Based Model Analysis}\label{sec:experimental_results:model_experiments:metric_based_model_analysis}
Using the expression results from the previous experiments and the extracted age, gender, and race attributes, we evaluated the bias of all the models with the metrics introduced in Sec.~\ref{sec:literature_review:bias_in_fer_algorithms}.

The first metric, Equalized Odds, examines the fairness of True Positive (TP) and False Positive (FP) rates across different groups. Table~\ref{tbl:bias_equalized_odds} highlights that Neutral and Happy were the most biased expressions. Almost all maximum biases for demographic attributes originate from these two expressions. For the age attribute, Surprise emerged as the least biased expression, whereas for the gender and race attributes, Fear showed the lowest bias. The Max column of this table reveals that the age attribute consistently exhibited the lowest bias across all models. The Mean and STD columns further emphasize that, on average, race was the most biased attribute across all models, while age demonstrated the least bias. When comparing the Mean and STD values among the models, the Equalized Odds metric indicates that the ViT model was the most biased, while ResNet and XceptionNet demonstrated the lowest bias scores.

\begin{table}[t!] 
\caption{Bias analysis using the Equal Opportunity metric, across models and three demographic attributes: age, gender, and race. Bias is calculated for each expression, followed by the calculation of maximum, mean, and standard deviation for each attribute. Range is [0, 100].}
\label{tbl:bias_equal_opportunity}
\centering
\small
\setlength{\tabcolsep}{3pt} 
\resizebox{0.5\textwidth}{!}
{
\begin{tabular}{ll>{\color{mediumgray}}c>{\color{mediumgray}}c>{\color{mediumgray}}c>{\color{mediumgray}}c>{\color{mediumgray}}c>{\color{mediumgray}}c>{\color{mediumgray}}ccc}
\toprule
                                                    &        & \color{black}\textbf{Neu} & \color{black}\textbf{Hap} & \color{black}\textbf{Sad} 
                                                             & \color{black}\textbf{Sur} & \color{black}\textbf{Fea} & \color{black}\textbf{Dis} 
                                                             & \color{black}\textbf{Ang} & \textbf{Max} & \textbf{Mean ± STD} \\
\midrule
\multicolumn{1}{l}{}                               & Age    & 8.3 & 2.0 & 4.0 & 5.2 & 4.6 & 0.4 & 1.5 & 8.3 & ~3.7 ± 2.4 \\
\multicolumn{1}{l}{}                               & Gender & 1.8 & 11.8 & 4.4 & 4.4 & 1.1 & 7.7 & 4.6 & 11.8 & ~5.1 ± 3.3 \\
\multicolumn{1}{l}{\multirow{-3}{*}{MobileNet}}    & Race   & 3.2 & 16.5 & 17.9 & 20.5 & 6.8 & 14.4 & 4.8 & 20.5 & 12.0 ± 6.4 \\ \midrule
\multicolumn{1}{l}{}                               & Age    & 9.5 & 3.0 & 1.8 & 4.9 & 3.8 & 2.3 & 0.7 & 9.5 & ~3.7 ± 2.6 \\
\multicolumn{1}{l}{}                               & Gender & 1.2 & 9.5 & 0.4 & 4.5 & 3.8 & 2.2 & 9.0 & 9.5 &~ 4.3 ± 3.3 \\
\multicolumn{1}{l}{\multirow{-3}{*}{ResNet}}       & Race   & 5.6 & 14.9 & 21.3 & 21.2 & 8.8 & 15.1 & 2.9 & 21.3 & 12.8 ± 6.7 \\ \midrule
\multicolumn{1}{l}{}                               & Age    & 8.5 & 4.3 & 0.6 & 6.8 & 6.0 & 1.3 & 3.1 & 8.5 & ~4.3 ± 2.6 \\
\multicolumn{1}{l}{}                               & Gender & 1.2 & 10.9 & 0.8 & 4.9 & 2.6 & 9.0 & 8.5 & 10.9 & ~5.4 ± 3.7 \\
\multicolumn{1}{l}{\multirow{-3}{*}{XceptionNet}}  & Race   & 6.1 & 15.7 & 16.1 & 19.2 & 10.4 & 11.0 & 5.7 & 19.2 & 12.0 ± 4.7 \\ \midrule
\multicolumn{1}{l}{}                               & Age    & 5.8 & 1.6 & 4.3 & 3.7 & 4.4 & 0.1 & 1.4 & 5.8 & ~3.0 ± 1.8 \\
\multicolumn{1}{l}{}                               & Gender & 4.3 & 11.9 & 7.1 & 8.3 & 3.6 & 5.5 & 11.0 & 11.9 & ~7.3 ± 2.9 \\
\multicolumn{1}{l}{\multirow{-3}{*}{ViT}}          & Race   & 4.6 & 17.1 & 24.9 & 25.2 & 11.1 & 9.6 & 10.4 & 25.2 & 14.7 ± 7.3 \\ \midrule
\multicolumn{1}{l}{}                               & Age    & 7.3 & 1.7 & 3.1 & 9.4 & 6.6 & 4.4 & 0.2 & 9.4 & ~4.6 ± 3.0 \\
\multicolumn{1}{l}{}                               & Gender & 2.8 & 9.0 & 2.8 & 7.9 & 7.4 & 7.4 & 15.4 & 15.4 & ~7.5 ± 3.9 \\
\multicolumn{1}{l}{\multirow{-3}{*}{CLIP}}         & Race   & 6.5 & 12.0 & 21.4 & 15.4 & 27.3 & 7.6 & 13.0 & 27.3 & 14.7 ± 6.8 \\ \midrule
\multicolumn{1}{l}{}                               & Age    & 5.6 & 0.5 & 0.4 & 5.0 & 1.9 & 1.3 & 0.2 & 5.6 & ~2.1 ± 2.0 \\
\multicolumn{1}{l}{}                               & Gender & 1.6 & 7.8 & 1.4 & 8.9 & 10.2 & 6.0 & 11.6 & 11.6 & ~6.7 ± 3.7 \\
\multicolumn{1}{l}{\multirow{-3}{*}{POSTER}}       & Race   & 6.0 & 8.2 & 12.1 & 24.3 & 25.0 & 19.2 & 8.0 & 25.0 & 14.6 ± 7.4 \\ \midrule
\multicolumn{1}{l}{}                               & Age    & 2.8 & 0.3 & 2.8 & 7.9 & 0.4 & 5.6 & 0.4 & 7.9 & ~2.8 ± 2.7 \\
\multicolumn{1}{l}{}                               & Gender & 5.3 & 6.8 & 1.7 & 5.9 & 8.5 & 7.2 & 11.6 & 11.6 & ~6.7 ± 2.8 \\
\multicolumn{1}{l}{\multirow{-3}{*}{CEPrompt}}     & Race   & 8.8 & 8.5 & 11.0 & 24.2 & 20.0 & 9.5 & 6.6 & 24.2 & 12.6 ± 6.1 \\ 
\bottomrule
\end{tabular}
}
\end{table}

Another metric we introduced in Sec.\ref{sec:methodology:model_analysis} is Equal Opportunity, which focuses on the ratio of TP predictions across different groups. The information about this metric is provided in Table~\ref{tbl:bias_equal_opportunity}. Unlike Equalized Odds, this table shows that the Happy expression was not the most biased in terms of Equal Opportunity. For the age attribute, Neutral ranked highest in bias, while for gender and race, Happy and Surprise expressions exhibited the highest bias scores, respectively. Examining the Max, Mean, and STD columns of this table reveals that race is the most bias-affected attribute across all models. Based on this metric, CLIP and ViT exhibited the highest bias, while ResNet and MobileNet achieved the lowest bias scores.

Demographic Parity analyzes the likelihood of any group receiving a particular outcome. The data in Table~\ref{tbl:bias_demographic_parity} shows that, compared to other bias metrics, Demographic Parity results in lower bias scores. For the age attribute, the highest bias was observed in the Neutral expression, while Happy, Sad, and Surprise had minimal effects. In contrast, Happy was the most biased expression for the gender attribute, while Fear and Sad showed the least bias. For the race attribute, Neutral and Happy were the primary sources of bias, while Fear exhibited the lowest bias across all models. Although the Max column highlights higher biases for the gender attribute, the Mean column underscores race as the most biased attribute. As with the other bias metrics, age remained the least biased attribute. The Mean and STD columns illustrate that for the age attribute, the CLIP model was the most biased model, while for gender and race, the ViT model was the most biased model. Overall, considering the Mean bias scores, the ViT model emerged as the most biased, while ResNet demonstrated the least bias.

\begin{table}[t!] 
\caption{Bias analysis using the Demographic Parity metric, across models and three demographic attributes: age, gender, and race. Bias is calculated for each expression, followed by the calculation of maximum, mean, and standard deviation for each attribute. Range is [0, 100].}
\label{tbl:bias_demographic_parity}
\centering
\small
\setlength{\tabcolsep}{3pt} 
\resizebox{0.489\textwidth}{!}
{
\begin{tabular}{ll>{\color{mediumgray}}c>{\color{mediumgray}}c>{\color{mediumgray}}c>{\color{mediumgray}}c>{\color{mediumgray}}c>{\color{mediumgray}}c>{\color{mediumgray}}ccc}
\toprule
                                                   &        & \color{black}\textbf{Neu} & \color{black}\textbf{Hap} & \color{black}\textbf{Sad} 
                                                            & \color{black}\textbf{Sur} & \color{black}\textbf{Fea} & \color{black}\textbf{Dis} 
                                                            & \color{black}\textbf{Ang} & \textbf{Max} & \textbf{Mean ± STD} \\
\midrule
\multicolumn{1}{l}{}                              & Age & 5.0 & 0.9 & 0.9 & 0.3 & 1.0 & 3.1 & 1.1 & 5.0 & ~1.7 ± 1.5 \\
\multicolumn{1}{l}{}                              & Gender & 8.4 & 15.6 & 0.7 & 2.3 & 0.1 & 7.4 & 2.9 & 15.6 & ~5.3 ± 5.1 \\
\multicolumn{1}{l}{\multirow{-3}{*}{MobileNet}}   & Race & 9.1 & 8.0 & 6.9 & 5.4 & 1.9 & 8.1 & 2.9 & 9.1 & ~6.0 ± 2.5 \\ \midrule
\multicolumn{1}{l}{}                              & Age & 6.5 & 0.8 & 0.8 & 0.1 & 1.7 & 1.0 & 2.2 & 6.5 & ~1.8 ± 1.9 \\
\multicolumn{1}{l}{}                              & Gender & 7.3 & 13.9 & 1.0 & 2.3 & 0.1 & 3.1 & 4.5 & 13.9 & ~4.6 ± 4.3 \\
\multicolumn{1}{l}{\multirow{-3}{*}{ResNet}}      & Race & 7.2 & 7.6 & 7.0 & 4.6 & 0.8 & 4.4 & 3.4 & 7.6 & ~5.0 ± 2.2 \\ \midrule
\multicolumn{1}{l}{}                              & Age & 5.1 & 0.3 & 1.3 & 0.4 & 2.0 & 1.3 & 1.2 & 5.1 & ~1.6 ± 1.5 \\
\multicolumn{1}{l}{}                              & Gender & 8.6 & 14.9 & 2.6 & 2.4 & 0.5 & 3.4 & 3.2 & 14.9 & ~5.0 ± 4.6 \\
\multicolumn{1}{l}{\multirow{-3}{*}{XceptionNet}} & Race & 10.0 & 9.5 & 6.8 & 5.4 & 1.6 & 4.8 & 2.6 & 10.0 & ~5.8 ± 2.9 \\ \midrule
\multicolumn{1}{l}{}                              & Age & 4.6 & 0.1 & 0.4 & 0.5 & 0.9 & 1.6 & 2.8 & 4.6 & ~1.5 ± 1.5 \\
\multicolumn{1}{l}{}                              & Gender & 11.3 & 16.1 & 0.1 & 2.8 & 0.3 & 3.1 & 5.0 & 16.1 & ~5.5 ± 5.5 \\
\multicolumn{1}{l}{\multirow{-3}{*}{ViT}}         & Race & 13.5 & 9.9 & 9.0 & 6.4 & 1.2 & 3.7 & 5.3 & 13.5 & ~7.0 ± 3.8 \\ \midrule
\multicolumn{1}{l}{}                              & Age & 6.5 & 1.8 & 0.2 & 1.1 & 2.7 & 0.1 & 2.9 & 6.5 & ~2.1 ± 2.0 \\
\multicolumn{1}{l}{}                              & Gender & 10.6 & 13.9 & 0.0 & 2.5 & 0.9 & 0.1 & 6.8 & 13.9 & ~4.9 ± 5.1 \\
\multicolumn{1}{l}{\multirow{-3}{*}{CLIP}}        & Race & 11.9 & 5.9 & 7.6 & 4.2 & 4.6 & 1.2 & 4.5 & 11.9 & ~5.7 ± 3.0 \\ \midrule
\multicolumn{1}{l}{}                              & Age & 6.0 & 0.7 & 0.6 & 0.7 & 2.0 & 0.3 & 3.1 & 6.0 & ~1.9 ± 1.9 \\
\multicolumn{1}{l}{}                              & Gender & 8.9 & 14.1 & 1.8 & 2.7 & 1.2 & 0.4 & 6.9 & 14.1 & ~5.1 ± 4.6 \\
\multicolumn{1}{l}{\multirow{-3}{*}{POSTER}}      & Race & 12.3 & 7.2 & 5.5 & 5.4 & 5.3 & 2.0 & 4.5 & 12.3 & ~6.0 ± 2.9 \\ \midrule
\multicolumn{1}{l}{}                              & Age & 4.9 & 0.7 & 0.2 & 0.9 & 1.8 & 0.2 & 3.2 & 4.9 & ~1.7 ± 1.6 \\
\multicolumn{1}{l}{}                              & Gender & 9.8 & 13.7 & 0.3 & 2.4 & 1.3 & 0.2 & 7.0 & 13.7 & ~4.9 ± 4.9 \\
\multicolumn{1}{l}{\multirow{-3}{*}{CEPrompt}}    & Race & 13.2 & 7.0 & 5.8 & 5.8 & 4.4 & 1.6 & 5.7 & 13.2 & ~6.2 ± 3.2 \\
\bottomrule
\end{tabular}
}
\end{table}

The final bias measure we examined is Treatment Equality, which assesses the distribution of misclassified samples across different groups. Table~\ref{tbl:bias_treatment_equality} demonstrates that Treatment Equality resulted in higher bias scores. A detailed expression-level analysis shows that Sad, Fear, and Disgust were the main sources of bias for the age attribute. Happy exhibited the highest bias for the gender and race attributes. Among the attributes, race had the highest bias scores, while age exhibited the lowest. This trend was consistent with our findings in the Equalized Odds and Equal Opportunity metrics. Comparing the Max column in Table~\ref{tbl:bias_treatment_equality} with the Max columns in Tables~\ref{tbl:bias_equalized_odds},~\ref{tbl:bias_equal_opportunity}, and~\ref{tbl:bias_demographic_parity} shows that Treatment Equality exhibited higher bias scores across all attributes. In a model-wise comparison of Treatment Equality, CLIP showed the lowest bias score, while the highest bias was observed in ViT.

To summarize, we analyzed model bias across individual attributes and expressions, followed by a comparative analysis of overall model-level bias. Our results show that ViT exhibits the highest level of bias under the Equalized Odds, Demographic Parity, and Treatment Equality metrics, while CLIP exhibits significant bias in Equal Opportunity. In contrast, ResNet is the most robust model against bias in three out of four bias metrics.

Investigating bias scores for FER-specialized models, POSTER~\cite{zheng2023poster} and CEPrompt~\cite{zhou2024ceprompt}, reveals that despite achieving high accuracies in FER task, they suffer from bias problem. Base on the ranking of models in each bias metric they are mainly located between the third to sixth place out of seven models, which is not considerable. A peer-to-peer comparison between POSTER and CEPrompt showed that in three out of four metrics CEPrompt showed lower bias than POSTER. 

\begin{table}[t!] 
\caption{Bias analysis using the Treatment Equality metric, across models and three demographic attributes: age, gender, and race. Bias is calculated for each expression, followed by the calculation of maximum, mean, and standard deviation for each attribute. Range is [0, 100].}
\label{tbl:bias_treatment_equality}
\centering
\small
\setlength{\tabcolsep}{3pt} 
\resizebox{0.5\textwidth}{!}
{
\begin{tabular}{ll>{\color{mediumgray}}c>{\color{mediumgray}}c>{\color{mediumgray}}c>{\color{mediumgray}}c>{\color{mediumgray}}c>{\color{mediumgray}}c>{\color{mediumgray}}ccc}
\toprule
                                                   &        & \color{black}\textbf{Neu} & \color{black}\textbf{Hap} & \color{black}\textbf{Sad} 
                                                            & \color{black}\textbf{Sur} & \color{black}\textbf{Fea} & \color{black}\textbf{Dis} 
                                                            & \color{black}\textbf{Ang} & \textbf{Max} & \textbf{Mean ± STD} \\
\midrule
\multicolumn{1}{l}{}                              & Age & 1.1 & 1.3 & 7.9 & 6.6 & 5.2 & 3.5 & 0.3 & 7.9 & ~3.7 ± 2.7\\
\multicolumn{1}{l}{}                              & Gender & 1.9 & 26.4 & 12.4 & 14.3 & 1.6 & 8.7 & 6.5 & 26.4 & 10.2 ± 7.9 \\
\multicolumn{1}{l}{\multirow{-3}{*}{MobileNet}}   & Race & 28.4 & 46.8 & 38.8 & 19.9 & 21.7 & 11.0 & 54.3 & 54.3 & ~31.5 ± 14.4 \\ \midrule
\multicolumn{1}{l}{}                              & Age & 8.4 & 3.3 & 11.0 & 11.3 & 14.1 & 13.6 & 5.1 & 14.1 & ~9.5 ± 3.8 \\
\multicolumn{1}{l}{}                              & Gender & 11.4 & 10.4 & 6.5 & 12.8 & 9.3 & 1.2 & 16.0 & 16.0 & ~9.6 ± 4.3 \\
\multicolumn{1}{l}{\multirow{-3}{*}{ResNet}}      & Race & 49.7 & 28.0 & 23.8 & 14.2 & 44.0 & 32.8 & 25.4 & 49.7 & ~31.1 ± 11.3 \\ \midrule
\multicolumn{1}{l}{}                              & Age & 1.5 & 11.4 & 0.6 & 1.7 & 14.5 & 17.3 & 1.5 & 17.3 & ~6.9 ± 6.6 \\
\multicolumn{1}{l}{}                              & Gender & 3.0 & 22.5 & 10.3 & 12.3 & 1.5 & 5.4 & 0.2 & 22.5 & ~7.8 ± 7.2 \\
\multicolumn{1}{l}{\multirow{-3}{*}{XceptionNet}} & Race & 35.7 & 48.2 & 14.5 & 24.9 & 33.9 & 29.8 & 19.1 & 48.2 & ~29.4 ± 10.4 \\ \midrule
\multicolumn{1}{l}{}                              & Age & 0.7 & 3.9 & 1.0 & 1.6 & 3.4 & 24.2 & 11.5 & 24.2 & ~6.6 ± 7.9 \\
\multicolumn{1}{l}{}                              & Gender & 7.7 & 34.8 & 2.8 & 4.9 & 1.0 & 16.0 & 16.2 & 34.8 & ~11.9 ± 10.8 \\
\multicolumn{1}{l}{\multirow{-3}{*}{ViT}}         & Race & 29.5 & 61.7 & 44.8 & 22.8 & 19.5 & 31.4 & 43.5 & 61.7 & ~36.1 ± 13.6 \\ \midrule
\multicolumn{1}{l}{}                              & Age & 6.5 & 8.3 & 6.3 & 2.0 & 23.5 & 0.0 & 1.8 & 23.5 & ~6.9 ± 7.3 \\
\multicolumn{1}{l}{}                              & Gender & 1.0 & 15.2 & 4.9 & 1.1 & 1.7 & 3.2 & 33.6 & 33.6 & ~~~8.6 ± 11.1 \\
\multicolumn{1}{l}{\multirow{-3}{*}{CLIP}}         & Race & 8.8 & 23.2 & 29.0 & 29.1 & 26.3 & 4.4 & 9.4 & 29.1 & 18.5 ± 9.8 \\ \midrule
\multicolumn{1}{l}{}                              & Age & 7.4 & 5.9 & 12.2 & 1.0 & 13.3 & 6.1 & 5.2 & 13.3 & ~7.3 ± 3.9 \\
\multicolumn{1}{l}{}                              & Gender & 1.4 & 31.9 & 20.5 & 6.5 & 6.4 & 0.5 & 14.3 & 31.9 & ~11.6 ± 10.5 \\
\multicolumn{1}{l}{\multirow{-3}{*}{POSTER}}      & Race & 10.7 & 37.1 & 24.3 & 30.7 & 37.5 & 12.2 & 39.4 & 39.4 & ~27.4 ± 11.1 \\ \midrule
\multicolumn{1}{l}{}                              & Age & 1.6 & 6.4 & 6.1 & 4.9 & 12.7 & 4.6 & 3.4 & 12.7 & ~5.6 ± 3.2 \\
\multicolumn{1}{l}{}                              & Gender & 8.7 & 23.1 & 6.3 & 17.8 & 8.3 & 8.6 & 21.7 & 23.1 & 13.5 ± 6.5 \\
\multicolumn{1}{l}{\multirow{-3}{*}{CEPrompt}}    & Race & 14.4 & 37.3 & 20.9 & 46.0 & 37.3 & 16.5 & 34.8 & 46.0 & ~29.6 ± 11.2 \\
\bottomrule
\end{tabular}
}
\end{table}

Table~\ref{tbl:bias_score_for_all_models} presents the final bias score of each model across different attributes and bias metrics. Based on Eq.~\ref{eq:bias_score_metric}, we used the maximum bias score of all the expressions for each bias metric, i.e., we used the Max column of Tables~\ref{tbl:bias_equalized_odds}-\ref{tbl:bias_treatment_equality} as the bias score of each metric. In the next step, using Eq.~\ref{eq:bias_score_attribute}, we calculated the average bias score for each model across all attributes (shown by Mean in Table~\ref{tbl:bias_score_for_all_models}). Finally, using Eq.~\ref{eq:bias_score}, the overall bias score for each model was obtained by averaging the bias scores for each attribute. 

\begin{table}[b!] 
\caption{Bias score for all models using the measures Equalized Odds (Eq-Od), Equal Opportunity (Eq-Op), Demographic Parity (De-Pa), and Treatment Equality (Tr-Eq) across age, race, and gender attributes. The mean represents the average of each row, and the bias score is the final mean score of the attributes. Data are in the range of [0, 100].}
\label{tbl:bias_score_for_all_models}
\centering
\small
\setlength{\tabcolsep}{5pt} 
\resizebox{1.0\linewidth}{!}
{
\begin{tabular}{lc>{\color{mediumgray}}c>{\color{mediumgray}}c>{\color{mediumgray}}c>{\color{mediumgray}}ccc}
\toprule
                                                   &        & \color{black}\textbf{Eq-Od} & \color{black}\textbf{Eq-Op} & \color{black}\textbf{De-Pa} 
                                                   & \color{black}\textbf{Tr-Eq} & \textbf{Mean} & \textbf{Bias Score}\\
\midrule
\multicolumn{1}{l}{}                              & Age    & 4.9 & 8.3 & 5.0 & 7.9 & 6.5 & \\
\multicolumn{1}{l}{}                              & Gender & 12.7 & 11.8 & 15.6 & 26.4 & 16.6 & \\
\multicolumn{1}{l}{\multirow{-3}{*}{MobileNet}}   & Race   & 11.6 & 20.5 & 9.1 & 54.3 & 23.8 & \multirow{-3}{*}{15.6} \\ \midrule
\multicolumn{1}{l}{}                              & Age    & 6.7 & 9.5 & 6.5 & 14.1 & 9.2  & \\
\multicolumn{1}{l}{}                              & Gender & 11.6 & 9.5 & 13.9 & 16.0 & 12.7 & \\
\multicolumn{1}{l}{\multirow{-3}{*}{ResNet}}      & Race   & 10.1 & 21.3 & 7.6 & 49.7 & 22.1 & \multirow{-3}{*}{\textbf{14.6}} \\ \midrule
\multicolumn{1}{l}{}                              & Age    & 5.1 & 8.5 & 5.1 & 17.3 & 9.0 & \\
\multicolumn{1}{l}{}                              & Gender & 12.2 & 10.9 & 14.9 & 22.5 & 15.1 & \\
\multicolumn{1}{l}{\multirow{-3}{*}{XceptionNet}} & Race   & 10.2 & 19.2 & 10.0 & 48.2 & 21.9 & \multirow{-3}{*}{15.3} \\ \midrule
\multicolumn{1}{l}{}                              & Age    & 4.7 & 5.8 & 4.6 & 24.2 & 9.8 & \\
\multicolumn{1}{l}{}                              & Gender & 13.4 & 11.9 & 16.1 & 34.8 & 19.0 & \\
\multicolumn{1}{l}{\multirow{-3}{*}{ViT}}         & Race   & 13.1 & 25.2 & 13.5 & 61.7 & 28.3 & \multirow{-3}{*}{19.0}  \\ \midrule
\multicolumn{1}{l}{}                              & Age    & 7.3 & 9.4 & 6.5 & 23.5 & 11.6 & \\
\multicolumn{1}{l}{}                              & Gender & 11.0 & 15.4 & 13.9 & 33.6 & 18.4 & \\
\multicolumn{1}{l}{\multirow{-3}{*}{CLIP}}         & Race   & 12.5 & 27.3 & 11.9 & 29.1 & 20.2 & \multirow{-3}{*}{16.7} \\ \midrule
\multicolumn{1}{l}{}                              & Age    & 6.4 & 5.6 & 6.0 & 13.3 & 7.8 & \\
\multicolumn{1}{l}{}                              & Gender & 10.6 & 11.6 & 14.1 & 31.9 & 17.0 & \\
\multicolumn{1}{l}{\multirow{-3}{*}{POSTER}}      & Race   & 10.8 & 25.0 & 12.3 & 39.4 & 21.8 & \multirow{-3}{*}{15.5} \\ \midrule
\multicolumn{1}{l}{}                              & Age    & 5.1 & 7.9 & 4.9 & 12.7 & 7.6 & \\
\multicolumn{1}{l}{}                              & Gender & 10.2 & 11.6 & 13.7 & 23.1 & 14.6 & \\
\multicolumn{1}{l}{\multirow{-3}{*}{CEPrompt}}    & Race   & 12.4 & 24.2 & 13.2 & 46.0 & 23.9 & \multirow{-3}{*}{15.4} \\
\bottomrule
\end{tabular}
}
\end{table}

The Mean column in this table reflected the average bias across fairness measures for each demographic attribute. Overall, race consistently exhibited the highest mean values across all models, indicating that race-related disparities are the most dominant source of bias in facial expression recognition. In contrast, age generally showed the lowest mean scores, suggesting relatively fair performance across age groups. Gender tended to fall between the two, with Transformer-based and large multimodal models, such as ViT and CLIP, showing the largest gender-related bias among all models.

In comparison, architectures such as ResNet and POSTER maintained more balanced mean values across demographics, implying steadier fairness behavior. Meanwhile, ViT displayed the largest disparities, reflecting greater sensitivity to demographic imbalances, particularly for gender and race. These findings emphasized that although recent models can reduce overall bias, they still struggle with demographic consistency.

The final bias scores showed that ViT was the most biased model, followed by CLIP in the second place. Additionally, this table highlights that residual-based models, including ResNet and XceptionNet, showed the minimum bias. The final bias score for the POSTER and CEPrompt is not considerable, where they ranked as the fourth and third least bias models. Ultimately, our experiments rank the models from the most biased to the least biased as follows: ViT, CLIP, MobileNet, POSTER, CEPrompt, XceptionNet, and ResNet. Notably, this research focused solely on the FER task, and the observed bias in the models could be evaluated on different tasks, potentially yielding varying bias scores.

\subsubsection{Conclusion}
\label{sec:experimental_results:model_experiments:conclusion}
Our findings reveal several key factors driving model bias in FER. First, CNN-based architectures exhibit bias primarily due to their reliance on localized texture and illumination cues that correlate with demographic attributes such as skin tone and lighting conditions. Second, Transformer-based models reduce this dependency through global attention mechanisms but still amplify demographic co-occurrence patterns embedded in the data, particularly across race. Third, FER-specialized models achieve higher accuracy, yet continue to reflect dataset-induced demographic correlations. Finally, the comparison across fairness metrics shows that model bias manifests not only in unequal accuracy but also in asymmetric error tendencies across groups. These results highlight that architectural improvements alone cannot eliminate bias. Fairness in FER depends on both model design and the diversity and balance of the training data.
\section{Discussion and Future Works} 
\label{sec:discussion_and_future_works}
The key contribution of this work demonstrates that traditional fairness metrics alone cannot fully characterize bias in FER models. By integrating dataset-level measures, multi-level co-occurrence analyses, and model fairness metrics, our study reveals how dataset imbalance and conditional dependencies align with and potentially propagate into architectural bias in CNNs, Transformers, and LVLMs. To the best of our knowledge, this joint dataset–model analysis has not been previously explored in FER and offers a methodological foundation for future fairness research in affective computing. In addition, our results highlight that fairness outcomes depend on the underlying metric and its multi-class adaptation. Binary fairness metrics provide complementary insights into model behavior when applied using a one-vs-rest formulation.

Our analysis highlights that both the datasets and models used for facial expression recognition exhibit significant biases that could impact the fairness and generalizability of FER systems. Across the four in-the-wild datasets of AffectNet, Fer2013, RAF-DB, and ExpW, demographic imbalances are apparent. Race consistently emerges as the most biased attribute, while age and gender exhibit moderate but persistent disparities. Similarly, expression distributions are skewed, with Neutral and Happy over-represented and Fear and Disgust underrepresented. These imbalances suggest that models trained on these datasets may inherit and even amplify existing biases.

The model evaluation further emphasizes the distinction between accuracy and fairness. While specialized FER models such as POSTER and CEPrompt achieve high overall accuracy, they do not effectively mitigate bias, particularly for underrepresented expressions and demographic groups. Despite strong performance, ViT was identified as the most biased model, whereas residual-based architectures like ResNet and XceptionNet demonstrated relatively lower bias. Interestingly, some models partially compensated for imbalances in the data, reducing the impact of skewed distributions, but no model completely addressed the fairness challenges inherent to FER tasks.

These findings underline two important points. First, high accuracy alone is not sufficient to ensure equitable FER performance across all demographic groups and expressions. Second, addressing bias requires attention at both the dataset and model levels: collecting more balanced datasets, particularly for underrepresented expressions and demographic groups, is essential, as is developing models that explicitly account for these biases during training. Overall, our results call for a holistic approach that evaluates FER systems not only on predictive performance but also on fairness metrics, ensuring a more reliable and equitable deployment in real-world applications.

Another important source of bias in FER lies in the annotation process itself. Facial expressions often contain inherent ambiguity, causing annotators to disagree due to cultural differences, subjective interpretation, and inconsistent labeling criteria. These factors introduce label noise and uncertainty that propagate into both dataset- and model-level bias, disproportionately affecting underrepresented demographic groups and low-frequency expressions. Recent studies suggest that soft-labels~\cite{fard2024affectnet+} or label-distribution learning can better capture the annotator uncertainty, allowing models to learn probabilistic expression representations rather than rigid categorical assignments, and semi-supervised frameworks that reduce dependence on fully-labeled data may further mitigate annotation-driven bias~\cite{zhang2025leaf}.

A further limitation concerns our reliance on an automated estimator (DeepFace) for demographic attributes. Although we apply the same estimator uniformly across all datasets and models, which preserves the validity of relative comparisons, automated demographic estimation can itself carry systematic errors that may correlate with the very attributes under study, particularly for race. Absolute bias magnitudes should therefore be interpreted with this caveat in mind, and future work would benefit from human-verified or self-reported demographic labels where available.

The contributions of this research are pivotal for advancing fairness in FER tasks. Below, we outline several open research directions in FER that can build on this study to address bias and promote fair decision-making:

\begin{itemize}
\item Developing more robust FER models by incorporating fairness-regularized loss functions during training. Future work could examine methods for implementing such constraints and their effects on model accuracy.
\item Designing debiased methods, such as adversarial training, balanced reweighting, or soft-label–based uncertainty modeling, to explicitly reduce demographic disparities and improve equitable FER performance.
\item Investigating fairness-aware training strategies, such as adversarial debiasing and reweighting, to mitigate demographic bias in FER models and analyze their impact on both accuracy and fairness.
\item Analyzing the evolution of bias throughout different stages of model development, including pretraining, training, fine-tuning, and domain adaptation.
\item Extending the use of our Conditional-Entropy Bias Index (CEBI) and Concentration Index (CI) in future studies to quantify dataset-level biases across broader affective computing and computer-vision applications.
\item Evaluating video-based datasets and models, where temporal dynamics between sequences are critical, to mitigate bias in video-based FER tasks.
\item Investigating bias in multi-modal models that leverage data from diverse sources such as images, text, video, audio, and physiological signals.
\item Exploring face generator models as potential sources of bias in future datasets and developing fair data generation methods to address these issues.
\item Investigating non-demographic sources of bias, such as illumination, background noise, head pose, gestures, eye gaze, and hair color, within datasets and their impact on fairness of models.
\item Reporting bias scores of the FER models alongside their accuracy, which enables a comprehensive evaluation, ensuring the effectiveness and fairness of the models.
\item Comparing the biases present in lab-controlled datasets versus in-the-wild datasets, and analyzing their influence on model training and fairness outcomes.
\end{itemize}

The methodological framework introduced in this study, including encompassing new dataset-bias metrics, multi-stage co-occurrence analysis, and cross-architecture fairness evaluation, provides a systematic foundation for future FER fairness research. By highlighting the need for balanced demographic distributions and fair model behavior, this unified benchmark offers a reference for assessing and improving fairness in emerging FER datasets and architectures. Such advances are essential for deploying FER systems in critical applications including human–computer interaction, mental health monitoring, and surveillance. Our newly developed CEBI and CI metrics provide deeper insight into dataset imbalance, suggesting future FER datasets may benefit from explicitly minimizing these two forms of statistical bias. Future research will build upon this framework to develop effective bias mitigation strategies while maintaining a high accuracy.

\appendix

\section{Theoretical Properties of CEBI and CI}
\label{app:appendix_a}

In this appendix, we discuss two newly introduced metrics, CEBI and CI to (i) identify the specific statistical object each metric operates on, (ii) prove a structural distinction between CEBI and GNMI that explains why they are not redundant despite occasional numerical similarity, (iii) do the same for CI relative to NSE, (iv) report a set of controlled synthetic cases that isolate each metric's blind spots, and (v) discuss the boundary condition of CEBI.

\subsection{Notation}
We retain the notation of Table~\ref{tbl:taxonomy}. For an attribute $A$ with groups
$a \in A$, group weight $p(A=a)$, label set $Y$ with $|Y|=n$, and
conditional entropy $H(Y\mid a) = H(Y \mid A=a)$, recall

\begin{equation}
\label{eq:appendix_a:conditional_entropy_bias_index}
\begin{adjustbox}{width=0.56\columnwidth}
$\begin{aligned}
CEBI = \frac{1}{k}\sum_{a\in A} \max\!\left(0,\; 1 - \frac{H(Y\mid a)}{H(Y)}\right),
\end{aligned}$
\end{adjustbox} 
\end{equation}

\begin{equation}
\label{eq:appendix_a:group_normalized_mutual_information}
\begin{adjustbox}{width=0.61\columnwidth}
$\begin{aligned}
GNMI(Y,A) = \frac{I(Y;A)}{\sqrt{H(Y)H(A)}}, \quad\quad\quad \\ 
I(Y;A) = \sum_{a \in A}\sum_{y \in Y}p(y,a) \log\frac{p(y, a)}{p(y)p(a)},\quad
\end{aligned}$
\end{adjustbox} 
\end{equation}

\begin{equation}
\label{eq:appendix_a:concentration_index}
\begin{adjustbox}{width=0.64\columnwidth}
$\begin{aligned}
CI =  \frac{1}{k} \sum_{a \in A} \frac{\left|\|\mathbf{p}\|_2^2 - \tfrac{1}{n}\right|}{1 - \tfrac{1}{n}}, \quad\|\mathbf{p}\|_2^2 = \sum_{y \in Y} p(y)^2,
\end{aligned}$
\end{adjustbox} 
\end{equation}

\begin{equation}
\label{eq:appendix_a:normalized_shannon_entropy}
\begin{adjustbox}{width=0.60\columnwidth}
$\begin{aligned}
NSE = \frac{1}{k} \sum_{a \in A} p(A=a) \left|1 - \frac{H(Y \mid a)}{\log n} \right|.
\end{aligned}$
\end{adjustbox} 
\end{equation}

\subsection{Statistical Taxonomy of Dataset-Bias Metrics}
\label{app:appendix_a:ststistical_taxonomy_of_dataset_bias_metrics}
Table~\ref{tbl:ststistical_taxonomy_of_dataset_bias_metrics} classifies each metric along four axes: (1) whether it operates on the joint, conditional, or marginal distribution; (2) whether it aggregates across groups (inter-group) or within a single group (intra-group); (3) whether the aggregation over groups is weighted by group prevalence $p(A=a)$ or is an unweighted mean over the $k$ groups; and (4) whether the metric is bounded or unbounded.

\begin{table}[t!]
\centering
\caption{Statistical taxonomy of the seven bias metrics. "Weighting"
refers to how each metric aggregates its per-group quantity into a
single score.}
\label{tbl:ststistical_taxonomy_of_dataset_bias_metrics}
\setlength{\tabcolsep}{6pt} 
\resizebox{1.0\linewidth}{!}
{
\renewcommand{\arraystretch}{1.4}
\begin{tabular}{lcccc}
\toprule
Metric  & Statistical object        & Locus                 & Weighting     & $[0,1]$ \\
\midrule
WD      & Marginal, pairwise        & Inter-group           & \xmark        & \cmark \\
JSD     & Marginal, pairwise        & Inter-group           & \xmark        & \cmark \\
NSE     & Conditional, per-group    & Intra-group           & \cmark        & \cmark \\
NLS     & Conditional, per-group    & Intra-group           & \xmark        & \cmark \\
GNMI    & Conditional / joint       & Inter-group           & \cmark        & \cmark \\
CEBI    & Conditional, per-group    & Inter-group$^\dagger$ & \xmark        & \cmark \\
CI      & Marginal, per-group       & Intra-group           & \xmark        & \cmark \\
\bottomrule
\end{tabular}
}
\\[1pt]
\raggedright\footnotesize $^\dagger$CEBI compares each group's
conditional entropy to the population entropy $H(Y)$, so although it
is computed per group it encodes an inter-group dependency signal,
distinguishing it from NSE and NLS, which are purely intra-group and
never reference $H(Y)$.
\end{table}

The two columns that matter most for the reviewers concerns are \emph{weighting} and \emph{locus}. GNMI is the only existing metric
that, like CEBI, targets conditional label--attribute dependency; CI's closest existing relative is NSE, which is also an intra-group
quantity. We study each pair in continue.

\subsection{CEBI Versus GNMI}
\label{app:appendix_a:cebi_versus_gnmi}

\textbf{Proposition 1:} \emph{GNMI is a population-weighted measure of label--attribute dependency, while CEBI is a group-size-invariant
measure of the same underlying quantity. Consequently, for any attribute in which a numerical minority group exhibits strong label--attribute dependency, GNMI can approach 0 while CEBI remains large, and the divergence between the two grows monotonically as the
minority group's prevalence $p(A=a) \to 0$.}

\textbf{Justification:} From Eq.~\ref{eq:appendix_a:group_normalized_mutual_information}, $I(Y;A) = H(Y) - \sum_a p(A=a) H(Y\mid a)$: the conditional entropy of every group is scaled by its prevalence before being subtracted from $H(Y)$. If a group $a^*$ has $H(Y\mid a^*) = 0$ (fully deterministic) but $p(A=a^*) = \epsilon$, its contribution to $I(Y;A)$ is bounded by $\epsilon \cdot H(Y)$ regardless of how extreme the dependency is inside that group, and GNMI $\to 0$ as $\epsilon \to 0$ even though group $a^*$ is maximally biased. By contrast, CEBI in Eq.~\ref{eq:appendix_a:conditional_entropy_bias_index} averages $1 - \frac{H(Y\mid a)}{H(Y)}$ over the $k$ groups with equal weight $\frac{1}{k}$, so a fully deterministic minority group contributes a full term of value $1$ to the average irrespective of $\epsilon$.

This difference is not just a mathematical property, but a deliberate design choice motivated by fairness research. Population-weighted measures, such as GNMI, are dominated by large demographic groups and may overlook severe bias affecting smaller groups. For this reason, many fairness methods also consider the worst-performing group rather than only the population average. CEBI follows the same idea for dataset bias analysis by asking whether any demographic subgroup exhibits strong label determinism, regardless of its size. In contrast, GNMI measures how much demographic information is conveyed by the labels of a randomly selected individual. Both perspectives are useful, but they answer different questions.

\textbf{Synthetic illustration:} We construct two binary-label ($n=2$) toy populations with two groups $a,b$ of prevalence
$p(A=a)=0.05$, $p(A=b)=0.95$, to isolate this effect.

\begin{itemize}
\item Case A (minority-driven dependency):
$p(y\mid a) = (1,0)$ (deterministic), $p(y\mid b) = (0.5,0.5)$
(uniform). Marginal $p(y) = (0.525, 0.475)$, $H(Y)=0.998$ bits.
\item Case B (majority-driven dependency):
The same two conditional distributions with roles swapped:
$p(y\mid a)=(0.5,0.5)$, $p(y\mid b)=(1,0)$. Marginal
$p(y)=(0.975,0.025)$, $H(Y)=0.169$ bits.
\end{itemize}

\begin{table}[t!]
\centering
\caption{GNMI and CEBI on two synthetic label distributions with dependency concentrated in different demographic groups. CEBI's group-size-independent weighting makes it insensitive to whether the minority (Case A) or majority (Case B) group drives the dependency, while GNMI changes substantially between the two cases because it weights groups by their prevalence.}
\label{tbl:cebi_and_gnmi_comparison}
\setlength{\tabcolsep}{6pt} 
\resizebox{1.0\linewidth}{!}
{
\renewcommand{\arraystretch}{1.4}
\begin{tabular}{llcc}
\toprule
Case    & Distributions                     & GNMI  & CEBI \\
\midrule
A       & 95\% uniform, 5\% deterministic\quad\quad\quad\quad\quad   & 0.09  & 0.50 \\
B       & 5\% uniform, 95\% deterministic\quad\quad\quad\quad\quad   & 0.54  & 0.50 \\
\bottomrule
\end{tabular}
}
\end{table}

Table~\ref{tbl:cebi_and_gnmi_comparison} illustrates this difference. GNMI changes substantially between Case A and Case B ($0.09$ vs. $0.54$) simply because the dependency occurs in a different demographic group. In contrast, CEBI gives the same value $(0.50)$ in both cases because it is unaffected by whether the minority or majority group exhibits the dependency. In Case~B the uniform minority group yields a negative per-group term that the $\max(0,\cdot)$ clipping sets to zero, as discussed in~\ref{app:appendix_a:boundary_behavior_of_cebi}; the deterministic majority group contributes the remaining term of value~1. This shows that CEBI is not a rescaled version of GNMI. Instead, it measures a different property that is independent of group size. As a result, the two metrics can differ considerably when dataset bias is concentrated in a demographic minority, which is exactly the type of bias that dataset auditing aims to identify.

\subsection{CI Versus NSE}
\label{app:appendix_a:ci_versus_nse}

\textbf{Proposition 2:} \emph{CI and NSE both measure label imbalance within a demographic group, but they are based on different diversity measures and therefore respond differently to the same label distribution. NSE is derived from normalized Shannon entropy and is more sensitive to the number of classes and the presence of rare labels. In contrast, CI is based on $\lVert p \rVert_2^2$, which gives greater weight to the most frequent class. As a result, when one class dominates the distribution, CI typically reports a larger bias value than NSE. Therefore, although both metrics measure intra-group label imbalance, they are not interchangeable.}

\textbf{Justification:} NSE (Eq.~\ref{eq:appendix_a:normalized_shannon_entropy}) is based on Shannon entropy, $H(Y\mid a) = -\sum_y p(y\mid a)\log p(y\mid a)$, which considers contributions from all classes, including rare ones. Therefore, NSE is relatively sensitive to the number of classes and the presence of under-represented labels. In contrast, CI (Eq.~\ref{eq:appendix_a:concentration_index}) is based on $\lVert p \rVert_2^2 = \sum_y p(y\mid a)^2$, which gives greater weight to classes with high probability and less weight to rare classes. Consequently, CI is more strongly influenced by the dominant class in the distribution. This behavior is consistent with the general properties of diversity indices: measures based on higher-order statistics place more emphasis on dominant categories and less on rare ones. As a result, NSE and CI can produce different scores depending on the source of the imbalance. NSE is more sensitive when bias arises from many under-represented classes, whereas CI responds more strongly when a single class dominates the distribution.

\textbf{Synthetic illustration:} We construct two toy scenarios, each consisting of a single demographic group ($k=1$) with its own label distribution over $n=5$ classes, so that each comparison isolates the shape of one distribution at a time, without any additional variation coming from how the metric aggregates scores across multiple groups. The two scenarios have the same number of populated categories but differ in \emph{where} the imbalance is concentrated.

\begin{itemize}
\item Case C (dominance-driven imbalance): One class carries almost all the mass, and the remaining four classes are small but
present: $p(y) = (0.80, 0.05, 0.05, 0.05, 0.05)$.
\item Case D (dispersion-driven imbalance): The mass is unevenly split among the classes without
one clear majority: $p(y) = (0.40, 0.30, 0.15, 0.10, 0.05)$.
\end{itemize}

\begin{table}[t!]
\centering
\caption{NSE and CI on two synthetic label distributions with different imbalance structure but comparable overall unevenness.
CI's quadratic weighting makes it substantially more sensitive to the single-dominant-class case (Case D), while NSE treats the two
cases more similarly because it weights rare classes more heavily than CI does.}
\label{tbl:ci_and_nse_comparison}
\setlength{\tabcolsep}{6pt} 
\resizebox{1.0\linewidth}{!}
{
\renewcommand{\arraystretch}{1.4}
\begin{tabular}{llcc}
\toprule
Case    & Distributions                                     & NSE   & CI \\
\midrule
C       & One dominant class (0.80), four small classes     & 0.62  & 0.69 \\
D       & Mass spread across classes, no dominant class     & 0.55  & 0.28 \\
\bottomrule
\end{tabular}
}
\end{table}

Table~\ref{tbl:ci_and_nse_comparison} shows the predicted pattern: NSE rates Case C and Case D as fairly similar in bias ($0.62$ vs. $0.55$), since both distributions deviate from uniform to a comparable degree in entropy terms. CI, by contrast, rates Case C as much more biased than Case D ($0.69$ vs. $0.28$), because Case C is dominated by a single class while Case D's imbalance is spread out. This confirms that CI and NSE are not redundant restatements of one another: they can disagree substantially on which of two datasets is "more biased," depending on whether the imbalance takes the form of one dominant class or a more diffuse unevenness, and reporting both gives a more complete picture than either alone.

\subsection{Boundary Behavior of CEBI}
\label{app:appendix_a:boundary_behavior_of_cebi}

CEBI (Eq.~\ref{eq:appendix_a:conditional_entropy_bias_index}) compares a
\emph{group-level} conditional entropy $H(Y\mid a)$ against the
\emph{population-level} entropy $H(Y)$ for each group $a$. Without
the $\max(0,\cdot)$ operation, the underlying per-group term
$1 - \frac{H(Y\mid a)}{H(Y)}$ can become negative whenever
$H(Y\mid a) > H(Y)$, i.e., whenever a numerical minority group is
markedly \emph{more} diverse in its label distribution than the
population as a whole. This is a genuine theoretical property of the
raw conditional-entropy ratio, and it motivates the clipping
operation built into Eq.~\ref{eq:appendix_a:conditional_entropy_bias_index}: by
taking $\max(0,\cdot)$ of each per-group term, CEBI discards this
negative signal rather than letting an unusually diverse minority
group offset or distort the bias attributed to other, genuinely
concentrated groups.

This clipping, i.e. $max(0,.)$, is also sufficient to
guarantee $\text{CEBI} \in [0,1]$, with no additional normalization
required. In fact, since entropy is non-negative, $H(Y\mid a) \ge 0$
for every group $a$, so the unclipped ratio always satisfies
$1 - \frac{H(Y\mid a)}{H(Y)} \le 1$. The upper bound therefore holds
automatically, before clipping is even applied. The $\max(0,\cdot)$
operation in Eq.~\ref{eq:appendix_a:conditional_entropy_bias_index}
additionally enforces a lower bound of $0$ on each per-group term.
Consequently, every per-group term lies in $[0,1]$, and since CEBI
is an unweighted average of these $k$ terms, it follows immediately
that $\text{CEBI} \in [0,1]$.

\subsection{Relation to Model-Fairness}
\label{app:appendix_a:relation_to_model_fairness}

It is important to note that CEBI and CI measure bias in the \emph{dataset}, before any model is trained. In contrast, Equalized Odds, Equal Opportunity, Demographic Parity, and Treatment Equality measure bias in a model's predictions $\hat Y$, after training. These two groups of metrics measure different things and one cannot replace the other. A dataset can have low CEBI or CI scores (little dependence between labels and demographic groups, and little imbalance within groups) and still produce a model with poor Equalized Odds scores, if the model's architecture or training process introduces bias on its own, separate from the data. Likewise, a dataset with high CEBI or CI scores does not always lead to a biased model, for example if a bias-mitigation technique such as reweighting is used during training. This is exactly why our unified framework reports both dataset-level and model-level bias metrics together: so that bias coming from the dataset and bias coming from the model can be identified separately.

\section*{Acknowledgment}
The authors acknowledge the use of AI-assisted language editing tools to improve clarity and readability. These tools were used only for minor linguistic refinement. All scientific content, including methodology, experiments, and conclusions, was developed independently by the authors.

\bibliographystyle{elsarticle-harv} 
\setlength{\bibsep}{0pt}
\bibliography{ref.bib}
\end{document}